\documentstyle[epic,eepic]{article}

\makeatletter

\newlength\titlebox 
 \twocolumn 

\def\addcontentsline#1#2#3{}

\def\maketitle{\par
 \begingroup
   \def\thefootnote{\fnsymbol{footnote}}
   \def\@makefnmark{\hbox to 0pt{$^{\@thefnmark}$\hss}}
   \twocolumn[\@maketitle] \@thanks
 \endgroup
 \setcounter{footnote}{0}
 \let\maketitle\relax \let\@maketitle\relax
 \gdef\@thanks{}\gdef\@author{}\gdef\@title{}\let\thanks\relax}
\def\@maketitle{\vbox to \titlebox{\hsize\textwidth
 \linewidth\hsize \vskip 0.625in minus 0.125in \centering
 {\Large\bf \@title \par} \vskip 0.2in plus 1fil minus 0.1in
 {\def\and{\unskip\enspace{\rm and}\enspace}%
  \def\And{\end{tabular}\hss \egroup \hskip 1in plus 2fil 
           \hbox to 0pt\bgroup\hss \begin{tabular}[t]{c}\bf}%
  \def\AND{\end{tabular}\hss\egroup \hfil\hfil\egroup
	  \vskip 0.25in plus 1fil minus 0.125in
	   \hbox to \linewidth\bgroup\large \hfil\hfil
 	     \hbox to 0pt\bgroup\hss \begin{tabular}[t]{c}\bf}
  \hbox to \linewidth\bgroup\large \hfil\hfil
    \hbox to 0pt\bgroup\hss \begin{tabular}[t]{c}\bf\@author 
			    \end{tabular}\hss\egroup
    \hfil\hfil\egroup}
  \vskip 0.3in plus 2fil minus 0.1in
}}
\renewenvironment{abstract}{\centerline{\large\bf
 Abstract}\vspace{0.5ex}\begin{quote}}{\par\end{quote}\vskip 1ex}

\def\thebibliography#1{\section*{References}
  \global\def\@listi{\leftmargin\leftmargini
               \labelwidth\leftmargini \advance\labelwidth-\labelsep
               \topsep 1pt plus 2pt minus 1pt
               \parsep 0.25ex plus 1pt \itemsep 0.25ex plus 1pt}
  \list {[\arabic{enumi}]}{\settowidth\labelwidth{[#1]}\leftmargin\labelwidth
    \advance\leftmargin\labelsep\usecounter{enumi}}
    \def\newblock{\hskip .11em plus .33em minus -.07em}
    \sloppy
    \sfcode`\.=1000\relax}

\def\@up#1{\raise.2ex\hbox{#1}}

\def\@citex[#1]#2{\if@filesw\immediate\write\@auxout{\string\citation{#2}}\fi
  \def\@citea{}\@cite{\@for\@citeb:=#2\do
     {\@citea\def\@citea{; }\@ifundefined
       {b@\@citeb}{{\bf ?}\@warning
        {Citation `\@citeb' on page \thepage \space undefined}}%
 {\csname b@\@citeb\endcsname}}}{#1}}

\let\@internalcite\cite
\def\cite{\def\citename##1{##1, }\@internalcite}
\def\shortcite{\def\citename##1{}\@internalcite}
\def\newcite{\leavevmode\def\citename##1{{##1} (}\@internalciteb}

\def\@citexb[#1]#2{\if@filesw\immediate\write\@auxout{\string\citation{#2}}\fi
  \def\@citea{}\@newcite{\@for\@citeb:=#2\do
    {\@citea\def\@citea{;\penalty\@m\ }\@ifundefined
       {b@\@citeb}{{\bf ?}\@warning
       {Citation `\@citeb' on page \thepage \space undefined}}%
\hbox{\csname b@\@citeb\endcsname}}}{#1}}
\def\@internalciteb{\@ifnextchar [{\@tempswatrue\@citexb}{\@tempswafalse\@citexb[]}}

\def\@newcite#1#2{{#1\if@tempswa, #2\fi)}}

\def\@biblabel#1{\def\citename##1{##1}[#1]\hfill}

\def\@cite#1#2{({#1\if@tempswa , #2\fi})}

\def\thebibliography#1{\vskip\parskip%
\vskip\baselineskip%
\def\baselinestretch{1}%
\ifx\@currsize\normalsize\@normalsize\else\@currsize\fi%
\vskip-\parskip%
\vskip-\baselineskip%
\section*{References\@mkboth
 {References}{References}}\list
 {}{\setlength{\labelwidth}{0pt}\setlength{\leftmargin}{\parindent}
 \setlength{\itemindent}{-\parindent}}
 \def\newblock{\hskip .11em plus .33em minus -.07em}
 \sloppy\clubpenalty4000\widowpenalty4000
 \sfcode`\.=1000\relax}

\def\thesourcebibliography#1{\vskip\parskip%
\vskip\baselineskip%
\def\baselinestretch{1}%
\ifx\@currsize\normalsize\@normalsize\else\@currsize\fi%
\vskip-\parskip%
\vskip-\baselineskip%
\section*{Sources of Attested Examples\@mkboth
 {Sources of Attested Examples}{Sources of Attested Examples}}\list
 {}{\setlength{\labelwidth}{0pt}\setlength{\leftmargin}{\parindent}
 \setlength{\itemindent}{-\parindent}}
 \def\newblock{\hskip .11em plus .33em minus -.07em}
 \sloppy\clubpenalty4000\widowpenalty4000
 \sfcode`\.=1000\relax}

\def\@lbibitem[#1]#2{\item[]\if@filesw 
      { \def\protect##1{\string ##1\space}\immediate
        \write\@auxout{\string\bibcite{#2}{#1}}\fi\ignorespaces}}

\def\@bibitem#1{\item\if@filesw \immediate\write\@auxout
       {\string\bibcite{#1}{\the\c@enumi}}\fi\ignorespaces}

\def\section{\@startsection {section}{1}{\z@}{-2.0ex plus
    -0.5ex minus -.2ex}{1.5ex plus 0.3ex minus .2ex}{\large\bf\raggedright}}
\def\subsection{\@startsection{subsection}{2}{\z@}{-1.8ex plus
    -0.5ex minus -.2ex}{0.8ex plus .2ex}{\normalsize\bf\raggedright}}
\def\subsubsection{\@startsection{subsubsection}{3}{\z@}{1.5ex plus
   0.5ex minus .2ex}{0.5ex plus .2ex}{\normalsize\bf\raggedright}}
\def\paragraph{\@startsection{paragraph}{4}{\z@}{1.5ex plus
   0.5ex minus .2ex}{-1em}{\normalsize\bf}}
\def\subparagraph{\@startsection{subparagraph}{5}{\parindent}{1.5ex plus
   0.5ex minus .2ex}{-1em}{\normalsize\bf}}

\skip\footins 9pt plus 4pt minus 2pt
\def\footnoterule{\kern-3pt \hrule width 5pc \kern 2.6pt }
\labelwidth\leftmargini\advance\labelwidth-\labelsep \labelsep 5pt

\def\@listi{\leftmargin\leftmargini}
\def\@listii{\leftmargin\leftmarginii
   \labelwidth\leftmarginii\advance\labelwidth-\labelsep
   \topsep 2pt plus 1pt minus 0.5pt
   \parsep 1pt plus 0.5pt minus 0.5pt
   \itemsep \parsep}
\def\@listiii{\leftmargin\leftmarginiii
    \labelwidth\leftmarginiii\advance\labelwidth-\labelsep
    \topsep 1pt plus 0.5pt minus 0.5pt 
    \parsep \z@ \partopsep 0.5pt plus 0pt minus 0.5pt
    \itemsep \topsep}
\def\@listiv{\leftmargin\leftmarginiv
     \labelwidth\leftmarginiv\advance\labelwidth-\labelsep}
\def\@listv{\leftmargin\leftmarginv
     \labelwidth\leftmarginv\advance\labelwidth-\labelsep}
\def\@listvi{\leftmargin\leftmarginvi
     \labelwidth\leftmarginvi\advance\labelwidth-\labelsep}

\belowdisplayskip \abovedisplayskip
\def\@normalsize{\@setsize\normalsize{11pt}\xpt\@xpt}
\def\small{\@setsize\small{10pt}\ixpt\@ixpt}
\def\footnotesize{\@setsize\footnotesize{10pt}\ixpt\@ixpt}
\def\scriptsize{\@setsize\scriptsize{8pt}\viipt\@viipt}
\def\tiny{\@setsize\tiny{7pt}\vipt\@vipt}
\def\large{\@setsize\large{14pt}\xiipt\@xiipt}
\def\Large{\@setsize\Large{16pt}\xivpt\@xivpt}
\def\LARGE{\@setsize\LARGE{20pt}\xviipt\@xviipt}
\def\huge{\@setsize\huge{23pt}\xxpt\@xxpt}
\def\Huge{\@setsize\Huge{28pt}\xxvpt\@xxvpt}

\makeatother

\newlength{\hbwidth}
\newcommand{\hb}[1]{\settowidth{\hbwidth}{#1}\makebox[\hbwidth][c]{}}

\title{Context-Sensitive Measurement of Word Distance \\
by Adaptive Scaling of a Semantic Space}

\author{Hideki Kozima \and Akira Ito \\
Kansai Advanced Research Center, \\
Communications Research Laboratory \\
588-2 Iwaoka, Nishi-ku, Kobe 651-24, JAPAN \\
E-mail: {\tt \{xkozima,ai\}@crl.go.jp}}

\begin{document}

\maketitle


\begin{abstract}
The paper proposes a computationally feasible method for measuring
context-sensitive semantic distance between words.  The distance is
computed by adaptive scaling of a semantic space.  In the semantic
space, each word in the vocabulary $V$ is represented by a
multi-dimensional vector which is obtained from an English dictionary
through a principal component analysis.  Given a word set $C$ which
specifies a context for measuring word distance, each dimension of the
semantic space is scaled up or down according to the distribution of $C$
in the semantic space.  In the space thus transformed, distance between
words in $V$ becomes dependent on the context $C$.  An evaluation
through a word prediction task shows that the proposed measurement
successfully extracts the context of a text.
\end{abstract}


\section{Introduction}

Semantic distance (or similarity) between words is one of the basic
measurements used in many fields of natural language processing,
information retrieval, etc.  Word distance provides bottom-up
information for text understanding and generation, since it indicates
semantic relationships between words that form a coherent text structure
(Grosz \& Sidner 86; Mann \& Thompson 87); word distance also provides
basis for episode association (Schank 90), since it works as associative
links between episodes.

A number of methods for measuring semantic distance between words have
been proposed in the studies of psycholinguistics, computational
linguistics, etc.  One of the pioneering works in psycholinguistics is
the ``semantic differential'' (Osgood 52), which analyzes meaning of
words by means of psychological experiments on human subjects.  Recent
studies in computational linguistics proposed computationally feasible
methods for measuring semantic word distance.  For example, (Morris \&
Hirst 91) used Roget's thesaurus as knowledge base for determining
whether or not two words are semantically related; (Brown et al.~92)
classified a vocabulary into semantic classes according to co-occurrency
of words in large corpora; (Kozima \& Furugori 93) computed similarity
between words by means of spreading activation on a semantic network of
an English dictionary.

The measurements in the former studies above are so-called context-free
or static ones, since they measure word distance irrespective of
contexts.  However, word distance changes in different contexts.  For
example, from the word {\tt car}, we can associate related words in the
following two directions.
\begin{itemize}
\item {\tt car} $\rightarrow$ 
      {\tt bus}, {\tt taxi}, {\tt railway}, $\cdots$
\item {\tt car} $\rightarrow$ 
      {\tt engine}, {\tt tire}, {\tt seat}, $\cdots$
\end{itemize}
The former is in the context of ``vehicle'', and the latter in the
context of ``components of a car''.  Even in free-association tasks, we
often imagine a certain context for retrieving related words.

In this paper, we will incorporate the context-sensitivity into semantic
distance between words.  A context can be specified by a word set $C$
consisting of keywords of the context (for instance, $C$ $\!=\!$
\{{\tt car}, {\tt bus}\} for the context ``vehicle'').  Now we can
exemplify the context-sensitive word association as follows.
\begin{itemize}
\item $C\!=\!$ \{{\tt car}, {\tt bus}\}\\
      \hb{$C\!$} $\rightarrow$
      {\tt taxi}, {\tt railway}, {\tt airplane}, $\cdots$
\item $C\!=\!$ \{{\tt car}, {\tt engine}\}\\
      \hb{$C\!$} $\rightarrow$
      {\tt tire}, {\tt seat}, {\tt headlight}, $\cdots$
\end{itemize}
Generally, if we change the context $C$, we will observe different
distance for the same word pair.  So, we in this paper will deal with
the following problem.
\begin{quote}
Under the context specified by a given word set $C$, compute semantic
distance $d(w, w'|C)$ between any two words $w, w'$ in our vocabulary
$V$.
\end{quote}

Our strategy for computing the context-sensitive word distance is
``adaptive scaling of a semantic space''.  Section 2 introduces the
semantic space where each word in the vocabulary $V$ is represented by a
multi-dimensional semantic vector.  The semantic vectors, called
Q-vectors, are obtained through a principal component analysis on
P-vectors.  P-vectors are generated by spreading activation on the
semantic network which is systematically constructed from an English
dictionary.  Section 3 describes adaptive scaling of the semantic space.
For a given word set $C$ that specifies a context, each dimension of the
semantic space is scaled up or down according to the distribution of $C$
in the semantic space.  In the semantic space thus transformed, distance
between Q-vectors becomes dependent on the given context.  Section 4
shows some examples of the context-sensitive word distance computed by
this method.  Section 5 evaluates the proposed measurement through word
prediction, i.e.~predicting succeeding words by using preceding words in
a text.  Section 6 discusses some theoretical aspects of the proposed
method, and Section 7 gives conclusion of this paper and puts this work
in perspective.


\section{Vector-Representation of Word Meaning}

Each word in the vocabulary $V$ is represented by a multi-dimensional
Q-vector.  In order to obtain Q-vectors, we first generate
2851-dimensional P-vectors by spreading activation on a semantic network
of an English dictionary (Kozima \& Furugori 93).  Next, through a
principal component analysis on P-vectors, we map each P-vector onto a
Q-vector with a reduced number of dimensions. (See {\bf Figure 1}.)

\begin{figure}
\begin{center}\small
\setlength{\unitlength}{0.011in}
\begin{picture}(228,225)
\thinlines
\path(70,215)(70,195)(60,205)
\path(70,195)(80,205)
\path(70,125)(80,135)
\path(70,145)(70,125)(60,135)
\path(70,85)(80,95)
\path(70,105)(70,85)(60,95)
\path(70,15)(80,25)
\path(70,35)(70,15)(60,25)
\path(140,35)(140,85)(0,85)(0,35)(140,35)
\path(140,145)(140,195)(0,195)(0,145)(140,145)
\dottedline{2}(170,170)(140,170)(150,180)
\dottedline{2}(140,170)(150,160)
\dottedline{2}(170,60)(140,60)(150,70)
\dottedline{2}(140,60)(150,50)
\put(70,220){\makebox(0,0)[b]{word: $w \in V$}}
\put(70,111){\makebox(0,0)[b]{P-vector: $P(w)$}}
\put(70,  1){\makebox(0,0)[b]{Q-vector: $Q(w)$}}
\put(20,160){\makebox(0,0)[lb]{on semantic network}}
\put(20,175){\makebox(0,0)[lb]{spreading activation}}
\put(35,65){\makebox(0,0)[lb]{transformation}}
\put(10,50){\makebox(0,0)[lb]{by principal components}}
\put(175,76){\makebox(0,0)[lb]{principal}}
\put(175,64){\makebox(0,0)[lb]{component}}
\put(175,52){\makebox(0,0)[lb]{analysis}}
\put(175,39){\makebox(0,0)[lb]{on P-vectors}}
\put(175,180){\makebox(0,0)[lb]{English}}
\put(175,168){\makebox(0,0)[lb]{dictionary}}
\put(175,155){\makebox(0,0)[lb]{(LDOCE)}}
\end{picture}
\end{center}
\vspace*{2mm}
\centerline{{\bf Figure 1: }Mapping words onto Q-vectors}
\end{figure}


\subsection{From an English Dictionary to P-Vectors}

Every word $w$ in the vocabulary $V$ is mapped onto a P-vector $P(w)$ by
spreading activation on the semantic network.  The network is
systematically constructed from a subset of the English dictionary,
LDOCE ({\it Longman Dictionary of Contemporary English\/}).  The network
has 2851 nodes corresponding to the words in LDV ({\it Longman Defining
Vocabulary\/}, 2851 words).  The network has 295914 links between the
nodes --- each node has a set of links corresponding to the words in its
definition in LDOCE.  Note that every headword in LDOCE is defined by
using LDV only.  The network becomes a closed cross-reference network of
English words.

\begin{figure}[tb]
\begin{center}
\setlength{\unitlength}{0.95mm}
\begin{picture}(75, 65)
\thicklines
\drawline( 10.0000,  65.0000)( 10.0000,  34.0000)
\drawline( 10.0000,  34.0000)( 23.0000,   8.0000)
\drawline( 23.0000,   8.0000)( 75.0000,   8.0000)
\thinlines
\put( 30.8000,   7.0000){\makebox(0,0)[t]{\scriptsize 2}}
\put( 41.2000,   7.0000){\makebox(0,0)[t]{\scriptsize 4}}
\put( 51.6000,   7.0000){\makebox(0,0)[t]{\scriptsize 6}}
\put( 62.0000,   7.0000){\makebox(0,0)[t]{\scriptsize 8}}
\put( 72.4000,   7.0000){\makebox(0,0)[t]{\scriptsize 10}}
\drawline( 10.0000,  40.2000)(  9.0000,  40.2000)
\put(  8.0000,  40.2000){\makebox(0,0)[rt]{\scriptsize      0.2}}
\drawline( 10.0000,  46.4000)(  9.0000,  46.4000)
\put(  8.0000,  46.4000){\makebox(0,0)[rt]{\scriptsize      0.4}}
\drawline( 10.0000,  52.6000)(  9.0000,  52.6000)
\put(  8.0000,  52.6000){\makebox(0,0)[rt]{\scriptsize      0.6}}
\drawline( 10.0000,  58.8000)(  9.0000,  58.8000)
\put(  8.0000,  58.8000){\makebox(0,0)[rt]{\scriptsize      0.8}}
\drawline( 10.0000,  65.0000)(  9.0000,  65.0000)
\put(  8.0000,  65.0000){\makebox(0,0)[rt]{\scriptsize      1.0}}
\put( 10.0000,  32.7000){\makebox(0,0)[r]{\footnotesize\tt red\_2}}
\put( 11.3000,  30.1000){\makebox(0,0)[r]{\footnotesize\tt red\_1}}
\put( 12.6000,  27.5000){\makebox(0,0)[r]{\footnotesize\tt orange\_1}}
\put( 13.9000,  24.9000){\makebox(0,0)[r]{\footnotesize\tt pink\_1}}
\put( 15.2000,  22.3000){\makebox(0,0)[r]{\footnotesize\tt pink\_2}}
\put( 16.5000,  19.7000){\makebox(0,0)[r]{\footnotesize\tt blood\_1}}
\put( 17.8000,  17.1000){\makebox(0,0)[r]{\footnotesize\tt copper\_1}}
\put( 19.1000,  14.5000){\makebox(0,0)[r]{\footnotesize\tt purple\_1}}
\put( 20.4000,  11.9000){\makebox(0,0)[r]{\footnotesize\tt purple\_2}}
\put( 21.7000,   9.3000){\makebox(0,0)[r]{\footnotesize\tt rose\_2}}
\put(  3.0000,  61.0000){\makebox(0,0)[r]{\small }}
\put( 75.0000,   4.5000){\makebox(0,0)[rt]{\small $T$ (steps)}}
\drawline( 21.7000,  10.6000)( 26.9000,  10.6000)( 28.2000,   8.0000)( 23.0000,   8.0000)( 21.7000,  10.6000)
\drawline( 26.9000,  10.6000)( 32.1000,  10.6000)( 33.4000,   8.0000)( 28.2000,   8.0000)
\drawline( 32.1000,  10.6000)( 32.1000,  11.9996)( 37.3000,  11.9996)( 38.6000,   9.3996)( 33.4000,   9.3996)( 32.1000,  11.9996)
\drawline( 33.4000,   8.0000)( 33.4000,   9.3996)
\drawline( 38.6000,   9.3996)( 38.6000,   8.0000)
\drawline( 37.3000,  11.9996)( 37.3000,  12.4970)( 42.5000,  12.4970)( 43.8000,   9.8970)( 38.6000,   9.8970)( 37.3000,  12.4970)
\drawline( 38.6000,   8.0000)( 38.6000,   9.8970)
\drawline( 43.8000,   9.8970)( 43.8000,   8.0000)
\drawline( 42.5000,  12.4970)( 42.5000,  13.0713)( 47.7000,  13.0713)( 49.0000,  10.4713)( 43.8000,  10.4713)( 42.5000,  13.0713)
\drawline( 43.8000,   9.8970)( 43.8000,  10.4713)
\drawline( 49.0000,  10.4713)( 49.0000,   8.0000)
\drawline( 47.7000,  13.0713)( 47.7000,  13.3569)( 52.9000,  13.3569)( 54.2000,  10.7569)( 49.0000,  10.7569)( 47.7000,  13.3569)
\drawline( 49.0000,   8.0000)( 49.0000,  10.7569)
\drawline( 54.2000,  10.7569)( 54.2000,   8.0000)
\drawline( 52.9000,  13.3569)( 52.9000,  13.6020)( 58.1000,  13.6020)( 59.4000,  11.0020)( 54.2000,  11.0020)( 52.9000,  13.6020)
\drawline( 54.2000,   8.0000)( 54.2000,  11.0020)
\drawline( 59.4000,  11.0020)( 59.4000,   8.0000)
\drawline( 58.1000,  13.6020)( 58.1000,  13.7153)( 63.3000,  13.7153)( 64.6000,  11.1153)( 59.4000,  11.1153)( 58.1000,  13.7153)
\drawline( 59.4000,  11.0020)( 59.4000,  11.1153)
\drawline( 64.6000,  11.1153)( 64.6000,   8.0000)
\drawline( 63.3000,  13.7153)( 63.3000,  13.8272)( 68.5000,  13.8272)( 69.8000,  11.2272)( 64.6000,  11.2272)( 63.3000,  13.8272)
\drawline( 64.6000,   8.0000)( 64.6000,  11.2272)
\drawline( 69.8000,  11.2272)( 69.8000,   8.0000)
\drawline( 68.5000,  13.8272)( 68.5000,  13.8980)( 73.7000,  13.8980)( 75.0000,  11.2980)( 69.8000,  11.2980)( 68.5000,  13.8980)
\drawline( 69.8000,  11.2272)( 69.8000,  11.2980)
\drawline( 75.0000,  11.2980)( 75.0000,   8.0000)
\drawline( 20.4000,  13.2000)( 25.6000,  13.2000)( 26.9000,  10.6000)( 21.7000,  10.6000)( 20.4000,  13.2000)
\drawline( 25.6000,  13.2000)( 25.6000,  14.1172)( 30.8000,  14.1172)( 32.1000,  11.5172)( 26.9000,  11.5172)( 25.6000,  14.1172)
\drawline( 26.9000,  10.6000)( 26.9000,  11.5172)
\drawline( 32.1000,  11.5172)( 32.1000,  10.6000)
\drawline( 30.8000,  14.1172)( 30.8000,  14.5478)( 36.0000,  14.5478)( 37.2741,  11.9996)\drawline( 32.1000,  11.9478)( 32.1000,  11.9478)( 30.8000,  14.5478)
\drawline( 32.1000,  11.5172)( 32.1000,  11.9478)
\drawline( 36.0000,  14.5478)( 36.0000,  15.0561)( 41.2000,  15.0561)( 42.4795,  12.4970)\drawline( 37.3000,  12.4561)( 37.3000,  12.4561)( 36.0000,  15.0561)
\drawline( 37.3000,  11.9996)( 37.3000,  12.4561)
\drawline( 41.2000,  15.0561)( 41.2000,  15.4325)( 46.4000,  15.4325)( 47.5806,  13.0713)\drawline( 42.5000,  12.8325)( 41.2000,  15.4325)
\drawline( 42.5000,  12.4970)( 42.5000,  12.8325)
\drawline( 46.4000,  15.4325)( 46.4000,  15.7638)( 51.6000,  15.7638)( 52.8035,  13.3569)\drawline( 47.7000,  13.1638)( 47.7000,  13.1638)( 46.4000,  15.7638)
\drawline( 47.7000,  13.0713)( 47.7000,  13.1638)
\drawline( 51.6000,  15.7638)( 51.6000,  16.0018)( 56.8000,  16.0018)( 57.9999,  13.6020)\drawline( 52.9000,  13.4018)( 52.9000,  13.4018)( 51.6000,  16.0018)
\drawline( 52.9000,  13.3569)( 52.9000,  13.4018)
\drawline( 56.8000,  16.0018)( 56.8000,  16.2121)( 62.0000,  16.2121)( 63.2484,  13.7153)\drawline( 58.1000,  13.6121)( 58.1000,  13.6121)( 56.8000,  16.2121)
\drawline( 58.1000,  13.6020)( 58.1000,  13.6121)
\drawline( 62.0000,  16.2121)( 62.0000,  16.3785)( 67.2000,  16.3785)( 68.4757,  13.8272)\drawline( 63.3000,  13.7785)( 63.3000,  13.7785)( 62.0000,  16.3785)
\drawline( 63.3000,  13.7153)( 63.3000,  13.7785)
\drawline( 67.2000,  16.3785)( 67.2000,  16.5155)( 72.4000,  16.5155)( 73.7000,  13.9155)( 68.5000,  13.9155)( 67.2000,  16.5155)
\drawline( 68.5000,  13.8272)( 68.5000,  13.9155)
\drawline( 73.7000,  13.9155)( 73.7000,  13.8980)
\drawline( 19.1000,  15.8000)( 24.3000,  15.8000)( 25.6000,  13.2000)( 20.4000,  13.2000)( 19.1000,  15.8000)
\drawline( 24.3000,  15.8000)( 24.3000,  16.1879)( 29.5000,  16.1879)( 30.5354,  14.1172)\drawline( 25.6000,  13.5879)( 25.6000,  13.5879)( 24.3000,  16.1879)
\drawline( 25.6000,  13.2000)( 25.6000,  13.5879)
\drawline( 29.5000,  16.1879)( 29.5000,  16.7226)( 34.7000,  16.7226)( 35.7874,  14.5478)\drawline( 30.8000,  14.1226)( 29.5000,  16.7226)
\drawline( 34.7000,  16.7226)( 34.7000,  17.2505)( 39.9000,  17.2505)( 40.9972,  15.0561)\drawline( 36.0000,  14.6505)( 34.7000,  17.2505)
\drawline( 36.0000,  14.5478)( 36.0000,  14.6505)
\drawline( 39.9000,  17.2505)( 39.9000,  17.7336)( 45.1000,  17.7336)( 46.2506,  15.4325)\drawline( 41.2000,  15.1336)( 41.2000,  15.1336)( 39.9000,  17.7336)
\drawline( 41.2000,  15.0561)( 41.2000,  15.1336)
\drawline( 45.1000,  17.7336)( 45.1000,  18.1540)( 50.3000,  18.1540)( 51.4951,  15.7638)\drawline( 46.4000,  15.5540)( 45.1000,  18.1540)
\drawline( 46.4000,  15.4325)( 46.4000,  15.5540)
\drawline( 50.3000,  18.1540)( 50.3000,  18.4942)( 55.5000,  18.4942)( 56.7462,  16.0018)\drawline( 51.6000,  15.8942)( 50.3000,  18.4942)
\drawline( 51.6000,  15.7638)( 51.6000,  15.8942)
\drawline( 55.5000,  18.4942)( 55.5000,  18.7755)( 60.7000,  18.7755)( 61.9817,  16.2121)\drawline( 56.8000,  16.1755)( 55.5000,  18.7755)
\drawline( 56.8000,  16.0018)( 56.8000,  16.1755)
\drawline( 60.7000,  18.7755)( 60.7000,  19.0075)( 65.9000,  19.0075)( 67.2000,  16.4075)( 62.0000,  16.4075)( 60.7000,  19.0075)
\drawline( 62.0000,  16.2121)( 62.0000,  16.4075)
\drawline( 67.2000,  16.4075)( 67.2000,  16.3785)
\drawline( 65.9000,  19.0075)( 65.9000,  19.1976)( 71.1000,  19.1976)( 72.4000,  16.5976)( 67.2000,  16.5976)( 65.9000,  19.1976)
\drawline( 67.2000,  16.3785)( 67.2000,  16.5976)
\drawline( 72.4000,  16.5976)( 72.4000,  16.5155)
\drawline( 17.8000,  18.4000)( 23.0000,  18.4000)( 24.3000,  15.8000)\drawline( 19.1000,  15.8000)( 17.8000,  18.4000)
\drawline( 23.0000,  18.4000)( 23.0000,  21.3128)( 28.2000,  21.3128)( 29.5000,  18.7128)( 24.3000,  18.7128)( 23.0000,  21.3128)
\drawline( 24.3000,  16.1879)( 24.3000,  18.7128)
\drawline( 29.5000,  18.7128)( 29.5000,  16.1879)
\drawline( 28.2000,  21.3128)( 28.2000,  22.1862)( 33.4000,  22.1862)( 34.7000,  19.5862)( 29.5000,  19.5862)( 28.2000,  22.1862)
\drawline( 29.5000,  16.1879)( 29.5000,  19.5862)
\drawline( 34.7000,  19.5862)( 34.7000,  16.7226)
\drawline( 33.4000,  22.1862)( 33.4000,  23.0320)( 38.6000,  23.0320)( 39.9000,  20.4320)( 34.7000,  20.4320)( 33.4000,  23.0320)
\drawline( 34.7000,  16.7226)( 34.7000,  20.4320)
\drawline( 39.9000,  20.4320)( 39.9000,  17.7336)
\drawline( 38.6000,  23.0320)( 38.6000,  23.4475)( 43.8000,  23.4475)( 45.1000,  20.8475)( 39.9000,  20.8475)( 38.6000,  23.4475)
\drawline( 39.9000,  20.4320)( 39.9000,  20.8475)
\drawline( 45.1000,  20.8475)( 45.1000,  17.7336)
\drawline( 43.8000,  23.4475)( 43.8000,  23.7858)( 49.0000,  23.7858)( 50.3000,  21.1858)( 45.1000,  21.1858)( 43.8000,  23.7858)
\drawline( 45.1000,  20.8475)( 45.1000,  21.1858)
\drawline( 50.3000,  21.1858)( 50.3000,  18.1540)
\drawline( 49.0000,  23.7858)( 49.0000,  23.8992)( 54.2000,  23.8992)( 55.5000,  21.2992)( 50.3000,  21.2992)( 49.0000,  23.8992)
\drawline( 50.3000,  18.1540)( 50.3000,  21.2992)
\drawline( 55.5000,  21.2992)( 55.5000,  18.4942)
\drawline( 54.2000,  23.8992)( 54.2000,  24.0553)( 59.4000,  24.0553)( 60.7000,  21.4553)( 55.5000,  21.4553)( 54.2000,  24.0553)
\drawline( 55.5000,  18.4942)( 55.5000,  21.4553)
\drawline( 60.7000,  21.4553)( 60.7000,  18.7755)
\drawline( 59.4000,  24.0553)( 59.4000,  24.1627)( 64.6000,  24.1627)( 65.9000,  21.5627)( 60.7000,  21.5627)( 59.4000,  24.1627)
\drawline( 60.7000,  18.7755)( 60.7000,  21.5627)
\drawline( 65.9000,  21.5627)( 65.9000,  19.0075)
\drawline( 64.6000,  24.1627)( 64.6000,  24.2511)( 69.8000,  24.2511)( 71.1000,  21.6511)( 65.9000,  21.6511)( 64.6000,  24.2511)
\drawline( 65.9000,  19.0075)( 65.9000,  21.6511)
\drawline( 71.1000,  21.6511)( 71.1000,  19.1976)
\drawline( 16.5000,  21.0000)( 21.7000,  21.0000)( 23.0000,  18.4000)( 17.8000,  18.4000)( 16.5000,  21.0000)
\drawline( 21.7000,  21.0000)( 21.7000,  24.2730)( 26.9000,  24.2730)( 28.2000,  21.6730)( 23.0000,  21.6730)( 21.7000,  24.2730)
\drawline( 23.0000,  18.4000)( 23.0000,  21.6730)
\drawline( 28.2000,  21.6730)( 28.2000,  21.3128)
\drawline( 26.9000,  24.2730)( 26.9000,  25.0695)( 32.1000,  25.0695)( 33.4000,  22.4695)( 28.2000,  22.4695)( 26.9000,  25.0695)
\drawline( 28.2000,  21.3128)( 28.2000,  22.4695)
\drawline( 33.4000,  22.4695)( 33.4000,  22.1862)
\drawline( 32.1000,  25.0695)( 32.1000,  26.1729)( 37.3000,  26.1729)( 38.6000,  23.5729)( 33.4000,  23.5729)( 32.1000,  26.1729)
\drawline( 33.4000,  22.1862)( 33.4000,  23.5729)
\drawline( 38.6000,  23.5729)( 38.6000,  23.0320)
\drawline( 37.3000,  26.1729)( 37.3000,  26.5908)( 42.5000,  26.5908)( 43.8000,  23.9908)( 38.6000,  23.9908)( 37.3000,  26.5908)
\drawline( 38.6000,  23.5729)( 38.6000,  23.9908)
\drawline( 43.8000,  23.9908)( 43.8000,  23.4475)
\drawline( 42.5000,  26.5908)( 42.5000,  26.9895)( 47.7000,  26.9895)( 49.0000,  24.3895)( 43.8000,  24.3895)( 42.5000,  26.9895)
\drawline( 43.8000,  23.4475)( 43.8000,  24.3895)
\drawline( 49.0000,  24.3895)( 49.0000,  23.7858)
\drawline( 47.7000,  26.9895)( 47.7000,  27.1153)( 52.9000,  27.1153)( 54.2000,  24.5153)( 49.0000,  24.5153)( 47.7000,  27.1153)
\drawline( 49.0000,  23.7858)( 49.0000,  24.5153)
\drawline( 54.2000,  24.5153)( 54.2000,  23.8992)
\drawline( 52.9000,  27.1153)( 52.9000,  27.2914)( 58.1000,  27.2914)( 59.4000,  24.6914)( 54.2000,  24.6914)( 52.9000,  27.2914)
\drawline( 54.2000,  24.5153)( 54.2000,  24.6914)
\drawline( 59.4000,  24.6914)( 59.4000,  24.0553)
\drawline( 58.1000,  27.2914)( 58.1000,  27.3927)( 63.3000,  27.3927)( 64.6000,  24.7927)( 59.4000,  24.7927)( 58.1000,  27.3927)
\drawline( 59.4000,  24.0553)( 59.4000,  24.7927)
\drawline( 64.6000,  24.7927)( 64.6000,  24.1627)
\drawline( 63.3000,  27.3927)( 63.3000,  27.4863)( 68.5000,  27.4863)( 69.8000,  24.8863)( 64.6000,  24.8863)( 63.3000,  27.4863)
\drawline( 64.6000,  24.1627)( 64.6000,  24.8863)
\drawline( 69.8000,  24.8863)( 69.8000,  24.2511)
\drawline( 15.2000,  23.6000)( 20.4000,  23.6000)( 21.7000,  21.0000)( 16.5000,  21.0000)( 15.2000,  23.6000)
\drawline( 20.4000,  23.6000)( 20.4000,  27.6866)( 25.6000,  27.6866)( 26.9000,  25.0866)( 21.7000,  25.0866)( 20.4000,  27.6866)
\drawline( 21.7000,  21.0000)( 21.7000,  25.0866)
\drawline( 26.9000,  25.0866)( 26.9000,  24.2730)
\drawline( 25.6000,  27.6866)( 25.6000,  29.0112)( 30.8000,  29.0112)( 32.1000,  26.4112)( 26.9000,  26.4112)( 25.6000,  29.0112)
\drawline( 26.9000,  24.2730)( 26.9000,  26.4112)
\drawline( 32.1000,  26.4112)( 32.1000,  25.0695)
\drawline( 30.8000,  29.0112)( 30.8000,  30.6375)( 36.0000,  30.6375)( 37.3000,  28.0375)( 32.1000,  28.0375)( 30.8000,  30.6375)
\drawline( 32.1000,  25.0695)( 32.1000,  28.0375)
\drawline( 37.3000,  28.0375)( 37.3000,  26.5908)
\drawline( 36.0000,  30.6375)( 36.0000,  31.3958)( 41.2000,  31.3958)( 42.5000,  28.7958)( 37.3000,  28.7958)( 36.0000,  31.3958)
\drawline( 37.3000,  28.0375)( 37.3000,  28.7958)
\drawline( 42.5000,  28.7958)( 42.5000,  26.5908)
\drawline( 41.2000,  31.3958)( 41.2000,  32.0653)( 46.4000,  32.0653)( 47.7000,  29.4653)( 42.5000,  29.4653)( 41.2000,  32.0653)
\drawline( 42.5000,  26.5908)( 42.5000,  29.4653)
\drawline( 47.7000,  29.4653)( 47.7000,  26.9895)
\drawline( 46.4000,  32.0653)( 46.4000,  32.3542)( 51.6000,  32.3542)( 52.9000,  29.7542)( 47.7000,  29.7542)( 46.4000,  32.3542)
\drawline( 47.7000,  26.9895)( 47.7000,  29.7542)
\drawline( 52.9000,  29.7542)( 52.9000,  27.2914)
\drawline( 51.6000,  32.3542)( 51.6000,  32.6562)( 56.8000,  32.6562)( 58.1000,  30.0562)( 52.9000,  30.0562)( 51.6000,  32.6562)
\drawline( 52.9000,  29.7542)( 52.9000,  30.0562)
\drawline( 58.1000,  30.0562)( 58.1000,  27.2914)
\drawline( 56.8000,  32.6562)( 56.8000,  32.8411)( 62.0000,  32.8411)( 63.3000,  30.2411)( 58.1000,  30.2411)( 56.8000,  32.8411)
\drawline( 58.1000,  27.2914)( 58.1000,  30.2411)
\drawline( 63.3000,  30.2411)( 63.3000,  27.3927)
\drawline( 62.0000,  32.8411)( 62.0000,  32.9979)( 67.2000,  32.9979)( 68.5000,  30.3979)( 63.3000,  30.3979)( 62.0000,  32.9979)
\drawline( 63.3000,  27.3927)( 63.3000,  30.3979)
\drawline( 68.5000,  30.3979)( 68.5000,  27.4863)
\drawline( 13.9000,  26.2000)( 19.1000,  26.2000)( 20.4000,  23.6000)\drawline( 15.2000,  23.6000)( 13.9000,  26.2000)
\drawline( 19.1000,  26.2000)( 19.1000,  30.8704)( 24.3000,  30.8704)( 25.6000,  28.2704)( 20.4000,  28.2704)( 19.1000,  30.8704)
\drawline( 20.4000,  27.6866)( 20.4000,  28.2704)
\drawline( 24.3000,  30.8704)( 24.3000,  32.3144)( 29.5000,  32.3144)( 30.8000,  29.7144)( 25.6000,  29.7144)( 24.3000,  32.3144)
\drawline( 25.6000,  28.2704)( 25.6000,  29.7144)
\drawline( 30.8000,  29.7144)( 30.8000,  29.0112)
\drawline( 29.5000,  32.3144)( 29.5000,  34.0740)( 34.7000,  34.0740)( 36.0000,  31.4740)( 30.8000,  31.4740)( 29.5000,  34.0740)
\drawline( 30.8000,  29.0112)( 30.8000,  31.4740)
\drawline( 36.0000,  31.4740)( 36.0000,  30.6375)
\drawline( 34.7000,  34.0740)( 34.7000,  34.8847)( 39.9000,  34.8847)( 41.2000,  32.2847)( 36.0000,  32.2847)( 34.7000,  34.8847)
\drawline( 36.0000,  30.6375)( 36.0000,  32.2847)
\drawline( 41.2000,  32.2847)( 41.2000,  31.3958)
\drawline( 39.9000,  34.8847)( 39.9000,  35.6012)( 45.1000,  35.6012)( 46.4000,  33.0012)( 41.2000,  33.0012)( 39.9000,  35.6012)
\drawline( 41.2000,  32.2847)( 41.2000,  33.0012)
\drawline( 46.4000,  33.0012)( 46.4000,  32.0653)
\drawline( 45.1000,  35.6012)( 45.1000,  35.9033)( 50.3000,  35.9033)( 51.6000,  33.3033)( 46.4000,  33.3033)( 45.1000,  35.9033)
\drawline( 46.4000,  32.0653)( 46.4000,  33.3033)
\drawline( 51.6000,  33.3033)( 51.6000,  32.3542)
\drawline( 50.3000,  35.9033)( 50.3000,  36.2246)( 55.5000,  36.2246)( 56.8000,  33.6246)( 51.6000,  33.6246)( 50.3000,  36.2246)
\drawline( 51.6000,  33.3033)( 51.6000,  33.6246)
\drawline( 56.8000,  33.6246)( 56.8000,  32.6562)
\drawline( 55.5000,  36.2246)( 55.5000,  36.4201)( 60.7000,  36.4201)( 62.0000,  33.8201)( 56.8000,  33.8201)( 55.5000,  36.4201)
\drawline( 56.8000,  32.6562)( 56.8000,  33.8201)
\drawline( 62.0000,  33.8201)( 62.0000,  32.8411)
\drawline( 60.7000,  36.4201)( 60.7000,  36.5861)( 65.9000,  36.5861)( 67.2000,  33.9861)( 62.0000,  33.9861)( 60.7000,  36.5861)
\drawline( 62.0000,  32.8411)( 62.0000,  33.9861)
\drawline( 67.2000,  33.9861)( 67.2000,  32.9979)
\drawline( 12.6000,  28.8000)( 17.8000,  28.8000)( 19.1000,  26.2000)( 13.9000,  26.2000)( 12.6000,  28.8000)
\drawline( 17.8000,  28.8000)( 17.8000,  31.6722)( 23.0000,  31.6722)( 23.4009,  30.8704)\drawline( 19.1000,  29.0722)( 17.8000,  31.6722)
\drawline( 19.1000,  26.2000)( 19.1000,  29.0722)
\drawline( 23.0000,  31.6722)( 23.0000,  32.9311)( 28.2000,  32.9311)( 28.5083,  32.3144)\drawline( 24.0304,  30.8704)( 23.0000,  32.9311)
\drawline( 28.2000,  32.9311)( 28.2000,  34.9940)( 33.4000,  34.9940)( 33.8600,  34.0740)\drawline( 29.5000,  32.3940)( 28.2000,  34.9940)
\drawline( 33.4000,  34.9940)( 33.4000,  36.6524)( 38.6000,  36.6524)( 39.4838,  34.8847)\drawline( 34.6892,  34.0740)( 33.4000,  36.6524)
\drawline( 38.6000,  36.6524)( 38.6000,  38.0251)( 43.8000,  38.0251)( 45.0119,  35.6012)\drawline( 39.9000,  35.4251)( 38.6000,  38.0251)
\drawline( 39.9000,  34.8847)( 39.9000,  35.4251)
\drawline( 43.8000,  38.0251)( 43.8000,  39.0094)( 49.0000,  39.0094)( 50.3000,  36.4094)( 45.1000,  36.4094)( 43.8000,  39.0094)
\drawline( 45.1000,  35.6012)( 45.1000,  36.4094)
\drawline( 50.3000,  36.4094)( 50.3000,  36.2246)
\drawline( 49.0000,  39.0094)( 49.0000,  39.8041)( 54.2000,  39.8041)( 55.5000,  37.2041)( 50.3000,  37.2041)( 49.0000,  39.8041)
\drawline( 50.3000,  36.4094)( 50.3000,  37.2041)
\drawline( 55.5000,  37.2041)( 55.5000,  36.2246)
\drawline( 54.2000,  39.8041)( 54.2000,  40.4244)( 59.4000,  40.4244)( 60.7000,  37.8244)( 55.5000,  37.8244)( 54.2000,  40.4244)
\drawline( 55.5000,  36.2246)( 55.5000,  37.8244)
\drawline( 60.7000,  37.8244)( 60.7000,  36.4201)
\drawline( 59.4000,  40.4244)( 59.4000,  40.9140)( 64.6000,  40.9140)( 65.9000,  38.3140)( 60.7000,  38.3140)( 59.4000,  40.9140)
\drawline( 60.7000,  36.4201)( 60.7000,  38.3140)
\drawline( 65.9000,  38.3140)( 65.9000,  36.5861)
\drawline( 11.3000,  31.4000)( 11.3000,  45.3766)( 16.5000,  45.3766)( 17.8000,  42.7766)( 12.6000,  42.7766)( 11.3000,  45.3766)
\drawline( 12.6000,  28.8000)( 12.6000,  42.7766)
\drawline( 17.8000,  42.7766)( 17.8000,  28.8000)
\drawline( 16.5000,  45.3766)( 16.5000,  47.2101)( 21.7000,  47.2101)( 23.0000,  44.6101)( 17.8000,  44.6101)( 16.5000,  47.2101)
\drawline( 17.8000,  28.8000)( 17.8000,  44.6101)
\drawline( 23.0000,  44.6101)( 23.0000,  31.6722)
\drawline( 21.7000,  47.2101)( 21.7000,  50.5084)( 26.9000,  50.5084)( 28.2000,  47.9084)( 23.0000,  47.9084)( 21.7000,  50.5084)
\drawline( 23.0000,  31.6722)( 23.0000,  47.9084)
\drawline( 28.2000,  47.9084)( 28.2000,  32.9311)
\drawline( 26.9000,  50.5084)( 26.9000,  51.7089)( 32.1000,  51.7089)( 33.4000,  49.1089)( 28.2000,  49.1089)( 26.9000,  51.7089)
\drawline( 28.2000,  32.9311)( 28.2000,  49.1089)
\drawline( 33.4000,  49.1089)( 33.4000,  36.6524)
\drawline( 32.1000,  51.7089)( 32.1000,  52.9073)( 37.3000,  52.9073)( 38.6000,  50.3073)( 33.4000,  50.3073)( 32.1000,  52.9073)
\drawline( 33.4000,  49.1089)( 33.4000,  50.3073)
\drawline( 38.6000,  50.3073)( 38.6000,  36.6524)
\drawline( 37.3000,  52.9073)( 37.3000,  53.0137)( 42.5000,  53.0137)( 43.8000,  50.4137)( 38.6000,  50.4137)( 37.3000,  53.0137)
\drawline( 38.6000,  50.3073)( 38.6000,  50.4137)
\drawline( 43.8000,  50.4137)( 43.8000,  38.0251)
\drawline( 42.5000,  53.0137)( 42.5000,  53.6096)( 47.7000,  53.6096)( 49.0000,  51.0096)( 43.8000,  51.0096)( 42.5000,  53.6096)
\drawline( 43.8000,  38.0251)( 43.8000,  51.0096)
\drawline( 49.0000,  51.0096)( 49.0000,  39.0094)
\drawline( 47.7000,  53.6096)( 47.7000,  53.9656)( 52.9000,  53.9656)( 54.2000,  51.3656)( 49.0000,  51.3656)( 47.7000,  53.9656)
\drawline( 49.0000,  51.0096)( 49.0000,  51.3656)
\drawline( 54.2000,  51.3656)( 54.2000,  39.8041)
\drawline( 52.9000,  53.9656)( 52.9000,  54.2759)( 58.1000,  54.2759)( 59.4000,  51.6759)( 54.2000,  51.6759)( 52.9000,  54.2759)
\drawline( 54.2000,  39.8041)( 54.2000,  51.6759)
\drawline( 59.4000,  51.6759)( 59.4000,  40.4244)
\drawline( 58.1000,  54.2759)( 58.1000,  54.5069)( 63.3000,  54.5069)( 64.6000,  51.9069)( 59.4000,  51.9069)( 58.1000,  54.5069)
\drawline( 59.4000,  40.4244)( 59.4000,  51.9069)
\drawline( 64.6000,  51.9069)( 64.6000,  40.9140)
\drawline( 10.0000,  34.0000)( 10.0000,  47.9766)( 15.2000,  47.9766)( 16.5000,  45.3766)( 11.3000,  45.3766)( 10.0000,  47.9766)
\drawline( 11.3000,  31.4000)( 11.3000,  45.3766)
\drawline( 16.5000,  45.3766)( 16.5000,  45.3766)
\drawline( 15.2000,  47.9766)( 15.2000,  54.4665)( 20.4000,  54.4665)( 21.7000,  51.8665)( 16.5000,  51.8665)( 15.2000,  54.4665)
\drawline( 16.5000,  45.3766)( 16.5000,  51.8665)
\drawline( 21.7000,  51.8665)( 21.7000,  47.2101)
\drawline( 20.4000,  54.4665)( 20.4000,  57.6108)( 25.6000,  57.6108)( 26.9000,  55.0108)( 21.7000,  55.0108)( 20.4000,  57.6108)
\drawline( 21.7000,  47.2101)( 21.7000,  55.0108)
\drawline( 26.9000,  55.0108)( 26.9000,  50.5084)
\drawline( 25.6000,  57.6108)( 25.6000,  59.4213)( 30.8000,  59.4213)( 32.1000,  56.8213)( 26.9000,  56.8213)( 25.6000,  59.4213)
\drawline( 26.9000,  50.5084)( 26.9000,  56.8213)
\drawline( 32.1000,  56.8213)( 32.1000,  51.7089)
\drawline( 30.8000,  59.4213)( 30.8000,  60.5308)( 36.0000,  60.5308)( 37.3000,  57.9308)( 32.1000,  57.9308)( 30.8000,  60.5308)
\drawline( 32.1000,  51.7089)( 32.1000,  57.9308)
\drawline( 37.3000,  57.9308)( 37.3000,  53.0137)
\drawline( 36.0000,  60.5308)( 36.0000,  61.2609)( 41.2000,  61.2609)( 42.5000,  58.6609)( 37.3000,  58.6609)( 36.0000,  61.2609)
\drawline( 37.3000,  57.9308)( 37.3000,  58.6609)
\drawline( 42.5000,  58.6609)( 42.5000,  53.0137)
\drawline( 41.2000,  61.2609)( 41.2000,  61.6668)( 46.4000,  61.6668)( 47.7000,  59.0668)( 42.5000,  59.0668)( 41.2000,  61.6668)
\drawline( 42.5000,  53.0137)( 42.5000,  59.0668)
\drawline( 47.7000,  59.0668)( 47.7000,  53.9656)
\drawline( 46.4000,  61.6668)( 46.4000,  61.9760)( 51.6000,  61.9760)( 52.9000,  59.3760)( 47.7000,  59.3760)( 46.4000,  61.9760)
\drawline( 47.7000,  59.0668)( 47.7000,  59.3760)
\drawline( 52.9000,  59.3760)( 52.9000,  53.9656)
\drawline( 51.6000,  61.9760)( 51.6000,  62.2123)( 56.8000,  62.2123)( 58.1000,  59.6123)( 52.9000,  59.6123)( 51.6000,  62.2123)
\drawline( 52.9000,  53.9656)( 52.9000,  59.6123)
\drawline( 58.1000,  59.6123)( 58.1000,  54.2759)
\drawline( 56.8000,  62.2123)( 56.8000,  62.3927)( 62.0000,  62.3927)( 63.3000,  59.7927)( 58.1000,  59.7927)( 56.8000,  62.3927)
\drawline( 58.1000,  54.2759)( 58.1000,  59.7927)
\drawline( 63.3000,  59.7927)( 63.3000,  54.5069)
\end{picture}
\end{center}
\vspace*{-2mm}
\centerline{{\bf Figure 2: }Spreading activation on the network}
\end{figure}

Each node of the network can hold activity, and it flows through the
links.  Hence, activating a node of the network for a certain period of
time causes the activity to spread over the network and forms a pattern
of activity distribution on it.  {\bf Figure 2} shows the pattern
generated by activating the node {\tt red}; the graph plots the activity
values of 10 dominant nodes at each step of time.  We empirically found
that the activated pattern reaches equilibrium approximately after 10
steps, whereas it will never reach the actual equilibrium.

The P-vector $P(w)$ of a word $w$ is the pattern of activity
distribution generated by activating the node corresponding to $w$.
$P(w)$ is a 2851-dimensional vector consisting of activity values of the
nodes at $T$ $\!=\!$ 10 as an approximation of the equilibrium.  $P(w)$
indicates how strongly each node of the network is semantically related
with $w$.

We in this paper define the vocabulary $V$ as LDV (2851 words) in order
to make our argument and experiments simple.  Although $V$ is not a
large vocabulary, it covers 83.07\% of 1006815 words of the LOB corpus
with the help of a morphological analysis.  In addition, $V$ can be
extended to the set of all headwords in LDOCE (more than 56000 words).
Obviously, an LDV word is directly mapped onto a P-vector by spreading
activation on the network; a non-LDV word can be mapped indirectly onto
a P-vector by activating a set of the words in its dictionary
definition.  (Recall that every headword in LDOCE is defined by using
LDV only.)

\begin{figure}[tb]
\begin{center}\small
\setlength{\unitlength}{0.55mm}
\begin{picture}(135.0000, 86.0000)
\thicklines
\drawline( 30.0000, 10.0000)(130.0000, 10.0000)
\put(130,0.5){\makebox(0,0)[r]{\small distance}}
\drawline( 30.0000, 10.0000)( 30.0000, 11.0000)
\put( 30.0000,  8.0000){\makebox(0,0)[t]{\scriptsize 0.0}}
\drawline( 46.6667, 10.0000)( 46.6667, 11.0000)
\put( 46.6667,  8.0000){\makebox(0,0)[t]{\scriptsize 0.2}}
\drawline( 63.3333, 10.0000)( 63.3333, 11.0000)
\put( 63.3333,  8.0000){\makebox(0,0)[t]{\scriptsize 0.4}}
\drawline( 80.0000, 10.0000)( 80.0000, 11.0000)
\put( 80.0000,  8.0000){\makebox(0,0)[t]{\scriptsize 0.6}}
\drawline( 96.6667, 10.0000)( 96.6667, 11.0000)
\put( 96.6667,  8.0000){\makebox(0,0)[t]{\scriptsize 0.8}}
\drawline(113.3333, 10.0000)(113.3333, 11.0000)
\put(113.3333,  8.0000){\makebox(0,0)[t]{\scriptsize 1.0}}
\drawline(130.0000, 10.0000)(130.0000, 11.0000)
\put(130.0000,  8.0000){\makebox(0,0)[t]{\scriptsize 1.2}}
\thinlines
\drawline( 30.0000,  68.1818)( 63.5037,  68.1818)
\drawline( 30.0000,  75.4545)( 63.5037,  75.4545)
\drawline( 63.5037,  68.1818)( 63.5037,  75.4545)
\drawline( 63.5037,  71.8182)(113.3721,  71.8182)
\drawline( 30.0000,  82.7273)(113.3721,  82.7273)
\drawline(113.3721,  71.8182)(113.3721,  82.7273)
\drawline( 30.0000,  46.3636)( 77.6914,  46.3636)
\drawline( 30.0000,  53.6364)( 77.6914,  53.6364)
\drawline( 77.6914,  46.3636)( 77.6914,  53.6364)
\drawline( 77.6914,  50.0000)( 84.8952,  50.0000)
\drawline( 30.0000,  60.9091)( 84.8952,  60.9091)
\drawline( 84.8952,  50.0000)( 84.8952,  60.9091)
\drawline( 30.0000,  17.2727)( 68.8526,  17.2727)
\drawline( 30.0000,  24.5455)( 68.8526,  24.5455)
\drawline( 68.8526,  17.2727)( 68.8526,  24.5455)
\drawline( 68.8526,  20.9091)(110.6498,  20.9091)
\drawline( 30.0000,  31.8182)(110.6498,  31.8182)
\drawline(110.6498,  20.9091)(110.6498,  31.8182)
\drawline(110.6498,  26.3636)(110.8187,  26.3636)
\drawline( 30.0000,  39.0909)(110.8187,  39.0909)
\drawline(110.8187,  26.3636)(110.8187,  39.0909)
\drawline( 84.8952,  55.4545)(114.0737,  55.4545)
\drawline(110.8187,  32.7273)(114.0737,  32.7273)
\drawline(114.0737,  55.4545)(114.0737,  32.7273)
\drawline(114.0737,  44.0909)(119.6549,  44.0909)
\drawline(113.3721,  77.2727)(119.6549,  77.2727)
\drawline(119.6549,  44.0909)(119.6549,  77.2727)
\put( 29.0000, 82.7273){\makebox(0,0)[r]{\tt tail\_1}}
\put( 29.0000, 75.4545){\makebox(0,0)[r]{\tt rat\_1}}
\put( 29.0000, 68.1818){\makebox(0,0)[r]{\tt mouse\_1}}
\put( 29.0000, 60.9091){\makebox(0,0)[r]{\tt tiger\_1}}
\put( 29.0000, 53.6364){\makebox(0,0)[r]{\tt lion\_1}}
\put( 29.0000, 46.3636){\makebox(0,0)[r]{\tt cat\_1}}
\put( 29.0000, 39.0909){\makebox(0,0)[r]{\tt pet\_1}}
\put( 29.0000, 31.8182){\makebox(0,0)[r]{\tt pet\_2}}
\put( 29.0000, 24.5455){\makebox(0,0)[r]{\tt rabbit\_1}}
\put( 29.0000, 17.2727){\makebox(0,0)[r]{\tt fur\_1}}
\end{picture}
\end{center}
\centerline{{\bf Figure 3: }Hierarchical clustering of P-vectors}
\end{figure}

The P-vector $P(w)$ represents the meaning of the word $w$ as its
relationships with other words in the vocabulary $V$.  Geometric
distance between two P-vectors $P(w)$ and $P(w')$ indicates semantic
distance between the words $w$ and $w'$.  {\bf Figure 3} shows a part of
the result of hierarchical clustering on P-vectors, using Euclidean
distance between centers of clusters.  The dendrogram reflects intuitive
semantic similarity between words: for instance, {\tt rat}/{\tt mouse},
{\tt tiger}/{\tt lion}/{\tt cat}, etc.  However, the similarity thus
observed is context-free and static, and the purpose of this paper is to
make it context-sensitive and dynamic.


\subsection{From P-Vectors to Q-Vectors}

Through a principal component analysis, we map every P-vector into a
Q-vector on which we will define context-sensitive distance later.  The
principal component analysis on P-vectors provides a series of 2851
principal components.  The most significant $m$ principal components
work as new orthogonal axes, that span $m$-dimensional vector space.  By
the $m$ principal components, every P-vector (with 2851 dimensions) is
mapped onto a Q-vector (with $m$ dimensions).  The value of $m$, which
will be determined later, is much smaller than 2851.  This means not
only compression of the semantic information, but also elimination of
the noise in P-vectors.

First, we compute the principal components $X_1,$ $X_2,$ $\!\cdots,\!$
$X_{2851}$ --- each of them is a 2851-dimensional vector --- under the
following conditions.
\begin{itemize}
\item  For any $X_i$, its norm $|X_i|$ is 1.
\item  For any $X_i, X_j$ ($i \neq j$), their inner product 
       $(X_i, X_j)$ is 0.
\item  The variance $v_i$ of P-vectors projected onto $X_i$ 
       is not smaller than any $v_j$ ($j > i$).
\end{itemize}
In other words, $X_1$ is the first principal component which has the
largest variance of P-vectors, and $X_2$ is the second principal
component which has the second-largest variance of P-vectors, and so on.
Consequently, the set of principal components $X_1,$ $X_2,$
$\!\cdots,\!$ $X_{2851}$ provides a new orthonormal coordinate system
for P-vectors.

\begin{figure}[tb]
\begin{center}
\setlength{\unitlength}{0.9mm}
\begin{picture}(70,66)
\thicklines
\drawline(  8.7500,  8.2500)(  8.7500, 66.0000)
( 70.0000, 66.0000)( 70.0000,  8.2500)(  8.7500,  8.2500)
\put( 70.0000,  1.0000){\makebox(0,0)[br]{\small $m$}}
\drawline(  8.7500,  8.2500)(  8.7500,  9.2500)
\put(  8.7500,  6.6000){\makebox(0,0)[t]{\scriptsize $0$}}
\drawline( 29.1667,  8.2500)( 29.1667,  9.2500)
\put( 29.1667,  6.6000){\makebox(0,0)[t]{\scriptsize $1000$}}
\drawline( 49.5833,  8.2500)( 49.5833,  9.2500)
\put( 49.5833,  6.6000){\makebox(0,0)[t]{\scriptsize $2000$}}
\drawline( 70.0000,  8.2500)( 70.0000,  9.2500)
\put( 70.0000,  6.6000){\makebox(0,0)[t]{\scriptsize $3000$}}
\put(  0.0000, 60.2250){\makebox(0,0)[l]{\small $\%$}}
\drawline(  8.7500,  8.2500)(  9.7500,  8.2500)
\put(  7.0000,  8.2500){\makebox(0,0)[r]{\scriptsize $0$}}
\drawline(  8.7500, 19.8000)(  9.7500, 19.8000)
\put(  7.0000, 19.8000){\makebox(0,0)[r]{\scriptsize $20$}}
\drawline(  8.7500, 31.3500)(  9.7500, 31.3500)
\put(  7.0000, 31.3500){\makebox(0,0)[r]{\scriptsize $40$}}
\drawline(  8.7500, 42.9000)(  9.7500, 42.9000)
\put(  7.0000, 42.9000){\makebox(0,0)[r]{\scriptsize $60$}}
\drawline(  8.7500, 54.4500)(  9.7500, 54.4500)
\put(  7.0000, 54.4500){\makebox(0,0)[r]{\scriptsize $80$}}
\drawline(  8.7500, 66.0000)(  9.7500, 66.0000)
\put(  7.0000, 66.0000){\makebox(0,0)[r]{\scriptsize $100$}}
\thinlines
\drawline(  8.7704, 10.2877)(  8.9746, 13.5659)
\drawline(  8.9746, 13.5659)(  9.1788, 15.7778)
\drawline(  9.1788, 15.7778)(  9.3829, 17.6295)
\drawline(  9.3829, 17.6295)(  9.5871, 19.2603)
\drawline(  9.5871, 19.2603)(  9.7912, 20.7437)
\drawline(  9.7912, 20.7437)(  9.9954, 22.0943)
\drawline(  9.9954, 22.0943)( 10.1996, 23.3415)
\drawline( 10.1996, 23.3415)( 10.4038, 24.5139)
\drawline( 10.4038, 24.5139)( 10.6079, 25.6027)
\drawline( 10.6079, 25.6027)( 10.8121, 26.6248)
\drawline( 10.8121, 26.6248)( 11.0162, 27.5865)
\drawline( 11.0162, 27.5865)( 11.2204, 28.4892)
\drawline( 11.2204, 28.4892)( 11.4246, 29.3477)
\drawline( 11.4246, 29.3477)( 11.6288, 30.1663)
\drawline( 11.6288, 30.1663)( 11.8329, 30.9445)
\drawline( 11.8329, 30.9445)( 12.0371, 31.6893)
\drawline( 12.0371, 31.6893)( 12.2413, 32.4013)
\drawline( 12.2413, 32.4013)( 12.4454, 33.0835)
\drawline( 12.4454, 33.0835)( 12.6496, 33.7370)
\drawline( 12.6496, 33.7370)( 12.8537, 34.3634)
\drawline( 12.8537, 34.3634)( 13.0579, 34.9661)
\drawline( 13.0579, 34.9661)( 13.2621, 35.5498)
\drawline( 13.2621, 35.5498)( 13.4662, 36.1136)
\drawline( 13.4662, 36.1136)( 13.6704, 36.6584)
\drawline( 13.6704, 36.6584)( 13.8746, 37.1832)
\drawline( 13.8746, 37.1832)( 14.0787, 37.6908)
\drawline( 14.0787, 37.6908)( 14.2829, 38.1829)
\drawline( 14.2829, 38.1829)( 14.4871, 38.6601)
\drawline( 14.4871, 38.6601)( 14.6913, 39.1232)
\drawline( 14.6913, 39.1232)( 14.8954, 39.5730)
\drawline( 14.8954, 39.5730)( 15.0996, 40.0107)
\drawline( 15.0996, 40.0107)( 15.3038, 40.4369)
\drawline( 15.3038, 40.4369)( 15.5079, 40.8507)
\drawline( 15.5079, 40.8507)( 15.7121, 41.2549)
\drawline( 15.7121, 41.2549)( 15.9162, 41.6484)
\drawline( 15.9162, 41.6484)( 16.1204, 42.0318)
\drawline( 16.1204, 42.0318)( 16.3246, 42.4048)
\drawline( 16.3246, 42.4048)( 16.5287, 42.7684)
\drawline( 16.5287, 42.7684)( 16.7329, 43.1227)
\drawline( 16.7329, 43.1227)( 16.9371, 43.4685)
\drawline( 16.9371, 43.4685)( 17.1412, 43.8065)
\drawline( 17.1412, 43.8065)( 17.3454, 44.1357)
\drawline( 17.3454, 44.1357)( 17.5496, 44.4571)
\drawline( 17.5496, 44.4571)( 17.7538, 44.7713)
\drawline( 17.7538, 44.7713)( 17.9579, 45.0784)
\drawline( 17.9579, 45.0784)( 18.1621, 45.3786)
\drawline( 18.1621, 45.3786)( 18.3663, 45.6716)
\drawline( 18.3663, 45.6716)( 18.5704, 45.9584)
\drawline( 18.5704, 45.9584)( 18.7746, 46.2396)
\drawline( 18.7746, 46.2396)( 18.9787, 46.5156)
\drawline( 18.9787, 46.5156)( 19.1829, 46.7861)
\drawline( 19.1829, 46.7861)( 19.3871, 47.0513)
\drawline( 19.3871, 47.0513)( 19.5913, 47.3118)
\drawline( 19.5913, 47.3118)( 19.7954, 47.5674)
\drawline( 19.7954, 47.5674)( 19.9996, 47.8188)
\drawline( 19.9996, 47.8188)( 20.2037, 48.0663)
\drawline( 20.2037, 48.0663)( 20.4079, 48.3107)
\drawline( 20.4079, 48.3107)( 20.6121, 48.5519)
\drawline( 20.6121, 48.5519)( 20.8163, 48.7897)
\drawline( 20.8163, 48.7897)( 21.0204, 49.0244)
\drawline( 21.0204, 49.0244)( 21.2246, 49.2558)
\drawline( 21.2246, 49.2558)( 21.4288, 49.4838)
\drawline( 21.4288, 49.4838)( 21.6329, 49.7087)
\drawline( 21.6329, 49.7087)( 21.8371, 49.9303)
\drawline( 21.8371, 49.9303)( 22.0412, 50.1487)
\drawline( 22.0412, 50.1487)( 22.2454, 50.3640)
\drawline( 22.2454, 50.3640)( 22.4496, 50.5758)
\drawline( 22.4496, 50.5758)( 22.6538, 50.7844)
\drawline( 22.6538, 50.7844)( 22.8579, 50.9899)
\drawline( 22.8579, 50.9899)( 23.0621, 51.1924)
\drawline( 23.0621, 51.1924)( 23.2662, 51.3918)
\drawline( 23.2662, 51.3918)( 23.4704, 51.5885)
\drawline( 23.4704, 51.5885)( 23.6746, 51.7820)
\drawline( 23.6746, 51.7820)( 23.8788, 51.9726)
\drawline( 23.8788, 51.9726)( 24.0829, 52.1605)
\drawline( 24.0829, 52.1605)( 24.2871, 52.3455)
\drawline( 24.2871, 52.3455)( 24.4913, 52.5280)
\drawline( 24.4913, 52.5280)( 24.6954, 52.7075)
\drawline( 24.6954, 52.7075)( 24.8996, 52.8843)
\drawline( 24.8996, 52.8843)( 25.1038, 53.0584)
\drawline( 25.1038, 53.0584)( 25.3079, 53.2300)
\drawline( 25.3079, 53.2300)( 25.5121, 53.3992)
\drawline( 25.5121, 53.3992)( 25.7162, 53.5663)
\drawline( 25.7162, 53.5663)( 25.9204, 53.7308)
\drawline( 25.9204, 53.7308)( 26.1246, 53.8929)
\drawline( 26.1246, 53.8929)( 26.3287, 54.0529)
\drawline( 26.3287, 54.0529)( 26.5329, 54.2107)
\drawline( 26.5329, 54.2107)( 26.7371, 54.3664)
\drawline( 26.7371, 54.3664)( 26.9413, 54.5201)
\drawline( 26.9413, 54.5201)( 27.1454, 54.6716)
\drawline( 27.1454, 54.6716)( 27.3496, 54.8210)
\drawline( 27.3496, 54.8210)( 27.5538, 54.9685)
\drawline( 27.5538, 54.9685)( 27.7579, 55.1140)
\drawline( 27.7579, 55.1140)( 27.9621, 55.2574)
\drawline( 27.9621, 55.2574)( 28.1663, 55.3991)
\drawline( 28.1663, 55.3991)( 28.3704, 55.5388)
\drawline( 28.3704, 55.5388)( 28.5746, 55.6766)
\drawline( 28.5746, 55.6766)( 28.7787, 55.8127)
\drawline( 28.7787, 55.8127)( 28.9829, 55.9471)
\drawline( 28.9829, 55.9471)( 29.1871, 56.0797)
\drawline( 29.1871, 56.0797)( 29.3912, 56.2106)
\drawline( 29.3912, 56.2106)( 29.5954, 56.3396)
\drawline( 29.5954, 56.3396)( 29.7996, 56.4671)
\drawline( 29.7996, 56.4671)( 30.0038, 56.5930)
\drawline( 30.0038, 56.5930)( 30.2079, 56.7174)
\drawline( 30.2079, 56.7174)( 30.4121, 56.8402)
\drawline( 30.4121, 56.8402)( 30.6163, 56.9614)
\drawline( 30.6163, 56.9614)( 30.8204, 57.0813)
\drawline( 30.8204, 57.0813)( 31.0246, 57.1996)
\drawline( 31.0246, 57.1996)( 31.2288, 57.3164)
\drawline( 31.2288, 57.3164)( 31.4329, 57.4319)
\drawline( 31.4329, 57.4319)( 31.6371, 57.5461)
\drawline( 31.6371, 57.5461)( 31.8412, 57.6590)
\drawline( 31.8412, 57.6590)( 32.0454, 57.7706)
\drawline( 32.0454, 57.7706)( 32.2496, 57.8810)
\drawline( 32.2496, 57.8810)( 32.4537, 57.9900)
\drawline( 32.4537, 57.9900)( 32.6579, 58.0980)
\drawline( 32.6579, 58.0980)( 32.8621, 58.2047)
\drawline( 32.8621, 58.2047)( 33.0662, 58.3102)
\drawline( 33.0662, 58.3102)( 33.2704, 58.4145)
\drawline( 33.2704, 58.4145)( 33.4746, 58.5178)
\drawline( 33.4746, 58.5178)( 33.6788, 58.6199)
\drawline( 33.6788, 58.6199)( 33.8829, 58.7208)
\drawline( 33.8829, 58.7208)( 34.0871, 58.8207)
\drawline( 34.0871, 58.8207)( 34.2913, 58.9195)
\drawline( 34.2913, 58.9195)( 34.4954, 59.0172)
\drawline( 34.4954, 59.0172)( 34.6996, 59.1139)
\drawline( 34.6996, 59.1139)( 34.9038, 59.2095)
\drawline( 34.9038, 59.2095)( 35.1079, 59.3040)
\drawline( 35.1079, 59.3040)( 35.3121, 59.3976)
\drawline( 35.3121, 59.3976)( 35.5162, 59.4901)
\drawline( 35.5162, 59.4901)( 35.7204, 59.5815)
\drawline( 35.7204, 59.5815)( 35.9246, 59.6720)
\drawline( 35.9246, 59.6720)( 36.1287, 59.7615)
\drawline( 36.1287, 59.7615)( 36.3329, 59.8500)
\drawline( 36.3329, 59.8500)( 36.5371, 59.9375)
\drawline( 36.5371, 59.9375)( 36.7413, 60.0239)
\drawline( 36.7413, 60.0239)( 36.9454, 60.1094)
\drawline( 36.9454, 60.1094)( 37.1496, 60.1940)
\drawline( 37.1496, 60.1940)( 37.3538, 60.2776)
\drawline( 37.3538, 60.2776)( 37.5579, 60.3603)
\drawline( 37.5579, 60.3603)( 37.7621, 60.4419)
\drawline( 37.7621, 60.4419)( 37.9663, 60.5227)
\drawline( 37.9663, 60.5227)( 38.1704, 60.6026)
\drawline( 38.1704, 60.6026)( 38.3746, 60.6815)
\drawline( 38.3746, 60.6815)( 38.5787, 60.7596)
\drawline( 38.5787, 60.7596)( 38.7829, 60.8370)
\drawline( 38.7829, 60.8370)( 38.9871, 60.9134)
\drawline( 38.9871, 60.9134)( 39.1912, 60.9889)
\drawline( 39.1912, 60.9889)( 39.3954, 61.0638)
\drawline( 39.3954, 61.0638)( 39.5996, 61.1377)
\drawline( 39.5996, 61.1377)( 39.8038, 61.2110)
\drawline( 39.8038, 61.2110)( 40.0079, 61.2834)
\drawline( 40.0079, 61.2834)( 40.2121, 61.3549)
\drawline( 40.2121, 61.3549)( 40.4163, 61.4257)
\drawline( 40.4163, 61.4257)( 40.6204, 61.4958)
\drawline( 40.6204, 61.4958)( 40.8246, 61.5651)
\drawline( 40.8246, 61.5651)( 41.0288, 61.6338)
\drawline( 41.0288, 61.6338)( 41.2329, 61.7017)
\drawline( 41.2329, 61.7017)( 41.4371, 61.7689)
\drawline( 41.4371, 61.7689)( 41.6412, 61.8354)
\drawline( 41.6412, 61.8354)( 41.8454, 61.9013)
\drawline( 41.8454, 61.9013)( 42.0496, 61.9663)
\drawline( 42.0496, 61.9663)( 42.2537, 62.0308)
\drawline( 42.2537, 62.0308)( 42.4579, 62.0946)
\drawline( 42.4579, 62.0946)( 42.6621, 62.1578)
\drawline( 42.6621, 62.1578)( 42.8663, 62.2203)
\drawline( 42.8663, 62.2203)( 43.0704, 62.2821)
\drawline( 43.0704, 62.2821)( 43.2746, 62.3432)
\drawline( 43.2746, 62.3432)( 43.4787, 62.4037)
\drawline( 43.4787, 62.4037)( 43.6829, 62.4636)
\drawline( 43.6829, 62.4636)( 43.8871, 62.5228)
\drawline( 43.8871, 62.5228)( 44.0913, 62.5813)
\drawline( 44.0913, 62.5813)( 44.2954, 62.6391)
\drawline( 44.2954, 62.6391)( 44.4996, 62.6965)
\drawline( 44.4996, 62.6965)( 44.7037, 62.7531)
\drawline( 44.7037, 62.7531)( 44.9079, 62.8092)
\drawline( 44.9079, 62.8092)( 45.1121, 62.8646)
\drawline( 45.1121, 62.8646)( 45.3162, 62.9196)
\drawline( 45.3162, 62.9196)( 45.5204, 62.9738)
\drawline( 45.5204, 62.9738)( 45.7246, 63.0275)
\drawline( 45.7246, 63.0275)( 45.9288, 63.0806)
\drawline( 45.9288, 63.0806)( 46.1329, 63.1332)
\drawline( 46.1329, 63.1332)( 46.3371, 63.1851)
\drawline( 46.3371, 63.1851)( 46.5412, 63.2364)
\drawline( 46.5412, 63.2364)( 46.7454, 63.2871)
\drawline( 46.7454, 63.2871)( 46.9496, 63.3373)
\drawline( 46.9496, 63.3373)( 47.1538, 63.3868)
\drawline( 47.1538, 63.3868)( 47.3579, 63.4357)
\drawline( 47.3579, 63.4357)( 47.5621, 63.4841)
\drawline( 47.5621, 63.4841)( 47.7662, 63.5320)
\drawline( 47.7662, 63.5320)( 47.9704, 63.5792)
\drawline( 47.9704, 63.5792)( 48.1746, 63.6259)
\drawline( 48.1746, 63.6259)( 48.3787, 63.6720)
\drawline( 48.3787, 63.6720)( 48.5829, 63.7176)
\drawline( 48.5829, 63.7176)( 48.7871, 63.7626)
\drawline( 48.7871, 63.7626)( 48.9913, 63.8072)
\drawline( 48.9913, 63.8072)( 49.1954, 63.8511)
\drawline( 49.1954, 63.8511)( 49.3996, 63.8946)
\drawline( 49.3996, 63.8946)( 49.6037, 63.9376)
\drawline( 49.6037, 63.9376)( 49.8079, 63.9801)
\drawline( 49.8079, 63.9801)( 50.0121, 64.0221)
\drawline( 50.0121, 64.0221)( 50.2163, 64.0635)
\drawline( 50.2163, 64.0635)( 50.4204, 64.1045)
\drawline( 50.4204, 64.1045)( 50.6246, 64.1451)
\drawline( 50.6246, 64.1451)( 50.8287, 64.1850)
\drawline( 50.8287, 64.1850)( 51.0329, 64.2246)
\drawline( 51.0329, 64.2246)( 51.2371, 64.2636)
\drawline( 51.2371, 64.2636)( 51.4412, 64.3022)
\drawline( 51.4412, 64.3022)( 51.6454, 64.3403)
\drawline( 51.6454, 64.3403)( 51.8496, 64.3780)
\drawline( 51.8496, 64.3780)( 52.0538, 64.4151)
\drawline( 52.0538, 64.4151)( 52.2579, 64.4517)
\drawline( 52.2579, 64.4517)( 52.4621, 64.4879)
\drawline( 52.4621, 64.4879)( 52.6662, 64.5236)
\drawline( 52.6662, 64.5236)( 52.8704, 64.5588)
\drawline( 52.8704, 64.5588)( 53.0746, 64.5936)
\drawline( 53.0746, 64.5936)( 53.2788, 64.6279)
\drawline( 53.2788, 64.6279)( 53.4829, 64.6618)
\drawline( 53.4829, 64.6618)( 53.6871, 64.6953)
\drawline( 53.6871, 64.6953)( 53.8912, 64.7283)
\drawline( 53.8912, 64.7283)( 54.0954, 64.7608)
\drawline( 54.0954, 64.7608)( 54.2996, 64.7929)
\drawline( 54.2996, 64.7929)( 54.5037, 64.8245)
\drawline( 54.5037, 64.8245)( 54.7079, 64.8557)
\drawline( 54.7079, 64.8557)( 54.9121, 64.8865)
\drawline( 54.9121, 64.8865)( 55.1163, 64.9169)
\drawline( 55.1163, 64.9169)( 55.3204, 64.9469)
\drawline( 55.3204, 64.9469)( 55.5246, 64.9764)
\drawline( 55.5246, 64.9764)( 55.7287, 65.0055)
\drawline( 55.7287, 65.0055)( 55.9329, 65.0344)
\drawline( 55.9329, 65.0344)( 56.1371, 65.0627)
\drawline( 56.1371, 65.0627)( 56.3413, 65.0907)
\drawline( 56.3413, 65.0907)( 56.5454, 65.1183)
\drawline( 56.5454, 65.1183)( 56.7496, 65.1454)
\drawline( 56.7496, 65.1454)( 56.9537, 65.1722)
\drawline( 56.9537, 65.1722)( 57.1579, 65.1986)
\drawline( 57.1579, 65.1986)( 57.3621, 65.2246)
\drawline( 57.3621, 65.2246)( 57.5662, 65.2502)
\drawline( 57.5662, 65.2502)( 57.7704, 65.2754)
\drawline( 57.7704, 65.2754)( 57.9746, 65.3002)
\drawline( 57.9746, 65.3002)( 58.1788, 65.3246)
\drawline( 58.1788, 65.3246)( 58.3829, 65.3486)
\drawline( 58.3829, 65.3486)( 58.5871, 65.3723)
\drawline( 58.5871, 65.3723)( 58.7912, 65.3955)
\drawline( 58.7912, 65.3955)( 58.9954, 65.4184)
\drawline( 58.9954, 65.4184)( 59.1996, 65.4409)
\drawline( 59.1996, 65.4409)( 59.4038, 65.4631)
\drawline( 59.4038, 65.4631)( 59.6079, 65.4849)
\drawline( 59.6079, 65.4849)( 59.8121, 65.5063)
\drawline( 59.8121, 65.5063)( 60.0162, 65.5273)
\drawline( 60.0162, 65.5273)( 60.2204, 65.5480)
\drawline( 60.2204, 65.5480)( 60.4246, 65.5683)
\drawline( 60.4246, 65.5683)( 60.6287, 65.5882)
\drawline( 60.6287, 65.5882)( 60.8329, 65.6078)
\drawline( 60.8329, 65.6078)( 61.0371, 65.6270)
\drawline( 61.0371, 65.6270)( 61.2413, 65.6458)
\drawline( 61.2413, 65.6458)( 61.4454, 65.6642)
\drawline( 61.4454, 65.6642)( 61.6496, 65.6823)
\drawline( 61.6496, 65.6823)( 61.8537, 65.7001)
\drawline( 61.8537, 65.7001)( 62.0579, 65.7174)
\drawline( 62.0579, 65.7174)( 62.2621, 65.7344)
\drawline( 62.2621, 65.7344)( 62.4663, 65.7510)
\drawline( 62.4663, 65.7510)( 62.6704, 65.7674)
\drawline( 62.6704, 65.7674)( 62.8746, 65.7834)
\drawline( 62.8746, 65.7834)( 63.0787, 65.7990)
\drawline( 63.0787, 65.7990)( 63.2829, 65.8143)
\drawline( 63.2829, 65.8143)( 63.4871, 65.8292)
\drawline( 63.4871, 65.8292)( 63.6912, 65.8436)
\drawline( 63.6912, 65.8436)( 63.8954, 65.8575)
\drawline( 63.8954, 65.8575)( 64.0996, 65.8709)
\drawline( 64.0996, 65.8709)( 64.3038, 65.8840)
\drawline( 64.3038, 65.8840)( 64.5079, 65.8965)
\drawline( 64.5079, 65.8965)( 64.7121, 65.9085)
\drawline( 64.7121, 65.9085)( 64.9162, 65.9201)
\drawline( 64.9162, 65.9201)( 65.1204, 65.9312)
\drawline( 65.1204, 65.9312)( 65.3246, 65.9417)
\drawline( 65.3246, 65.9417)( 65.5288, 65.9417)
\drawline( 65.5288, 65.9517)( 65.7329, 65.9517)
\drawline( 65.7329, 65.9611)( 65.9371, 65.9611)
\drawline( 65.9371, 65.9698)( 66.1412, 65.9698)
\drawline( 66.1412, 65.9778)( 66.3454, 65.9778)
\drawline( 66.3454, 65.9851)( 66.5496, 65.9851)
\drawline( 66.5496, 65.9916)( 66.7537, 65.9916)
\drawline( 66.7537, 65.9968)( 66.9579, 65.9968)
\end{picture}
\end{center}
\vspace*{-2mm}
\centerline{{\bf Figure 4: }Cumulation of $v_i$ (percentage)}
\end{figure}

Next, we pick up the first $m$ principal components $X_1,$ $X_2,$
$\!\cdots,\!$ $X_m$.  The principal components are in descending order
of their significance, because the variance $v_i$ indicates the amount
of information represented by $X_i$.  The cumulative variances
$\sum^m_{i=1}$ $\!v_i$ in {\bf Figure 4} shows that even a few hundred
axes can represent more than half of the total information of P-vectors.
The amount of information represented by Q-vectors increases with $m$.
However, for large $m$, each Q-vector would be isolated because of
overfitting --- a large number of parameters could not be estimated by a
small number of data.

We estimate the optimal number of dimensions of Q-vectors at $m$ $\!=\!$
281 by minimizing the proportion of noise remained in Q-vectors.  The
amount of the noise is estimated by $\sum_{w \in F} |Q(w)|$, where $F$
($\subset\!$ $V$) is a set of 210 function words --- determiners,
articles, prepositions, pronouns, and conjunctions.  We estimated the
proportion of noise for all $m$ $\!=\!$ 1, $\!\cdots,\!$ 2851 and
obtained the minimum for $m$ $\!=\!$ 281.  Hereafter, we will use a
281-dimensional semantic space.

Lastly, we map each P-vector $P(w)$ onto a 281-dimensional Q-vector
$Q(w)$.  The $i$-th component of $Q(w)$ is the projected value of $P(w)$
on the principal component $X_i$; the origin of $X_i$ is set to the
average of the projected values on it.  We can ignore the direction of
$X_i$, which determines the sign of projected values, since it has no
effect on distance between Q-vectors.


\section{Adaptive Scaling of the Semantic Space}

Adaptive scaling of the semantic space of Q-vectors provides
context-sensitive and dynamic distance between Q-vectors.  Simple
Euclidean distance between Q-vectors is not so different from that
between P-vectors illustrated in Figure 3; both are context-free and
static distance.  The adaptive scaling process transforms the semantic
space so as to make it adapt to a given context $C$.  In the semantic
space thus transformed, simple Euclidean distance between Q-vectors
becomes dependent on $C$.  (See {\bf Figure 5}.)

\begin{figure}
\begin{center}
\setlength{\unitlength}{0.0125in}
\begin{picture}(175,160)
\thinlines
\put(60,150){\makebox(0,0)[b]{semantic space}}
\put(60,138){\makebox(0,0)[b]{of Q-vectors}}
\put(60,125){\makebox(0,0)[b]{(context-free)}}
\path(60,118)(60,98)
\path(50,108)(60,98)(70,108)
\path(120,63)(120,98)(0,98)(0,63)(120,63)
\put(60,78){\makebox(0,0)[b]{adaptive scaling}}
\path(60,63)(60,43)
\path(50,53)(60,43)(70,53)
\put(60,30){\makebox(0,0)[b]{semantic space}}
\put(60,18){\makebox(0,0)[b]{of Q-vectors}}
\put(60,5){\makebox(0,0)[b]{(context-sensitive)}}
\dottedline{2}(120,80)(140,80)
\dottedline{2}(130,90)(120,80)(130,70)
\put(165,80){\makebox(0,0)[b]{context}}
\put(165,69){\makebox(0,0)[b]{$C$}}
\end{picture}
\end{center}
\vspace*{0.5mm}
\centerline{{\bf Figure 5: }Adaptive scaling (overview)}
\end{figure}

\subsection{Semantic Subspaces}

A subspace of the semantic space of Q-vectors works as a simple device
for semantic word clustering.  In a semantic subspace with the
dimensions appropriately selected, the Q-vectors of semantically related
words are expected to form a cluster.  The reasons for this are as
follows.
\begin{itemize}
\item  Semantically related words have similar P-vectors, as illustrated 
       in Figure 3.
\item  The dimensions of Q-vectors are extracted from the correlations 
       between P-vectors by means of the principal component analysis.
\end{itemize}

As an example of word clustering in the semantic subspaces, let us
consider the following 15 words.
\begin{quote}
  1.~{\tt after}, \ 2.~{\tt ago}, \ 3.~{\tt before}, \ 4.~{\tt
  bicycle}, \ 5.~{\tt bus}, \ 6.~{\tt car}, \ 7.~{\tt enjoy}, 
  \ 8.~{\tt former}, \ 9.~{\tt glad}, \ 10.~{\tt good}, \ 11.~{\tt 
  late}, \ 12.~{\tt pleasant}, \ 13.~{\tt railway}, \ 14.~{\tt 
  satisfaction}, \ 15.~{\tt vehicle}.
\end{quote}
We scattered these words on the subspace $X_2$ $\!\times\!$ $X_3$,
namely the plane spanned by the second and third dimensions of
Q-vectors.  As shown in {\bf Figure 6}, the words form three apparent
clusters, namely ``goodness'', ``vehicle'', and ``past''.

However, it is still difficult to select appropriate dimensions for
making a semantic cluster for given words.  In the example above, we
used only two dimensions; most semantic clusters need more dimensions to
be well-separated.  Moreover, each of the 2851 dimensions is just
selected or discarded; this ignores their strengths of contribution to
forming clusters.

\begin{figure}[tb]
\begin{center}
\setlength{\unitlength}{0.9mm}
\begin{picture}(70,66)
\thicklines
\drawline(  8.7500,  8.2500)(  8.7500, 66.0000)
( 70.0000, 66.0000)( 70.0000,  8.2500)(  8.7500,  8.2500)
\put( 70.0000,  1.0000){\makebox(0,0)[br]{\small $X_2$}}
\drawline(  8.7500,  8.2500)(  8.7500,  9.2500)
\put(  8.7500,  6.6000){\makebox(0,0)[t]{\scriptsize $-0.4$}}
\drawline( 17.5000,  8.2500)( 17.5000,  9.2500)
\put( 17.5000,  6.6000){\makebox(0,0)[t]{\scriptsize $-0.3$}}
\drawline( 26.2500,  8.2500)( 26.2500,  9.2500)
\put( 26.2500,  6.6000){\makebox(0,0)[t]{\scriptsize $-0.2$}}
\drawline( 35.0000,  8.2500)( 35.0000,  9.2500)
\put( 35.0000,  6.6000){\makebox(0,0)[t]{\scriptsize $-0.1$}}
\drawline( 43.7500,  8.2500)( 43.7500,  9.2500)
\put( 43.7500,  6.6000){\makebox(0,0)[t]{\scriptsize $0$}}
\drawline( 52.5000,  8.2500)( 52.5000,  9.2500)
\put( 52.5000,  6.6000){\makebox(0,0)[t]{\scriptsize $0.1$}}
\drawline( 61.2500,  8.2500)( 61.2500,  9.2500)
\put( 61.2500,  6.6000){\makebox(0,0)[t]{\scriptsize $0.2$}}
\drawline( 70.0000,  8.2500)( 70.0000,  9.2500)
\put( 70.0000,  6.6000){\makebox(0,0)[t]{\scriptsize $0.3$}}
\put(  0.0000, 61.1875){\makebox(0,0)[l]{\small $X_3$}}
\drawline(  8.7500,  8.2500)(  9.7500,  8.2500)
\put(  7.0000,  8.2500){\makebox(0,0)[r]{\scriptsize $-0.4$}}
\drawline(  8.7500, 17.8750)(  9.7500, 17.8750)
\put(  7.0000, 17.8750){\makebox(0,0)[r]{\scriptsize $-0.3$}}
\drawline(  8.7500, 27.5000)(  9.7500, 27.5000)
\put(  7.0000, 27.5000){\makebox(0,0)[r]{\scriptsize $-0.2$}}
\drawline(  8.7500, 37.1250)(  9.7500, 37.1250)
\put(  7.0000, 37.1250){\makebox(0,0)[r]{\scriptsize $-0.1$}}
\drawline(  8.7500, 46.7500)(  9.7500, 46.7500)
\put(  7.0000, 46.7500){\makebox(0,0)[r]{\scriptsize $0$}}
\drawline(  8.7500, 56.3750)(  9.7500, 56.3750)
\put(  7.0000, 56.3750){\makebox(0,0)[r]{\scriptsize $0.1$}}
\drawline(  8.7500, 66.0000)(  9.7500, 66.0000)
\put(  7.0000, 66.0000){\makebox(0,0)[r]{\scriptsize $0.2$}}
\thinlines
\put( 43.4415, 30.2881){\makebox(0,0){\footnotesize\tt after}}
\put( 42.9243, 21.1996){\makebox(0,0){\footnotesize\tt ago}}
\put( 49.5137, 26.0032){\makebox(0,0){\footnotesize\tt before}}
\put( 61.8558, 52.9377){\makebox(0,0){\footnotesize\tt bicycle}}
\put( 53.5033, 48.9121){\makebox(0,0){\footnotesize\tt bus}}
\put( 58.7145, 51.0775){\makebox(0,0){\footnotesize\tt car}}
\put( 22.8458, 57.3982){\makebox(0,0){\footnotesize\tt enjoy}}
\put( 45.1560, 27.5315){\makebox(0,0){\footnotesize\tt former}}
\put( 19.5151, 50.8473){\makebox(0,0){\footnotesize\tt glad}}
\put( 21.4460, 49.0000){\makebox(0,0){\footnotesize\tt good}}
\put( 48.0094, 24.3047){\makebox(0,0){\footnotesize\tt late}}
\put( 21.5735, 55.2321){\makebox(0,0){\footnotesize\tt pleasant}}
\put( 58.7536, 47.3903){\makebox(0,0){\footnotesize\tt railway}}
\put( 21.6961, 47.0971){\makebox(0,0){\footnotesize\tt satisfaction}}
\put( 56.7007, 52.2778){\makebox(0,0){\footnotesize\tt vehicle}}
\dottedline{0.5}( 43.7500,  8.2500)( 43.7500, 66.0000)
\dottedline{0.5}(  8.7500, 46.7500)( 70.0000, 46.7500)
\end{picture}
\end{center}
\vspace*{-2mm}
\centerline{{\bf Figure 6: }Clusters in the semantic subspace}
\end{figure}


\begin{figure}
\begin{center}
\setlength{\unitlength}{0.010in}
\begin{picture}(240,365)
\thinlines
\put(120,300){\ellipse{162}{42}}
\dottedline{2}(0,300)(240,300)
\dottedline{2}(120,360)(120,240)
\put(10,340){\makebox(0,0)[lb]{before}}
\put(10,325){\makebox(0,0)[lb]{scaling}}
\put(120,295){\makebox(0,0)[b]{\large$C$}}
\put(130,340){\makebox(0,0){\small$\bullet$}}
\put(220,290){\makebox(0,0){\small$\circ$}}
\dashline{2}(201,300)(161,100)
\dashline{2}(39,300)(79,100)
\thicklines
\drawline(115,230)(115,215)
\drawline(125,230)(125,215)
\drawline(110,220)(120,210)(130,220)
\thinlines
\put(120,100){\ellipse{82}{82}}
\dottedline{2}(0,100)(240,100)
\dottedline{2}(120,200)(120,30)
\put(10,140){\makebox(0,0)[lb]{after}}
\put(10,125){\makebox(0,0)[lb]{scaling}}
\put(120,95){\makebox(0,0)[b]{\large$C$}}
\put(125,180){\makebox(0,0){\small$\bullet$}}
\put(168,80){\makebox(0,0){\small$\circ$}}
\end{picture}
\end{center}
\vspace*{-8mm}
\centerline{{\bf Figure 7: }Scaling the semantic space}
\end{figure}

\subsection{Adaptive Scaling}

Adaptive scaling of the semantic space provides a weight for each
dimension in order to form a desired semantic cluster; the weights are
given by scaling factors of the dimensions.  The method make the
semantic space adapt to a given context $C$ in the following way.
\begin{quote}
  Each dimension of the semantic space is scaled up or down, so as to
  make the words in $C$ form a cluster in the semantic space.
\end{quote}
In the semantic space thus transformed, the distance between Q-vectors
will change with $C$.  For example, as illustrated in {\bf Figure 7},
when $C$ has oval-shaped (generally, hyper-elliptic) distribution in the
pre-scaling space, each dimension is scaled up or down so that $C$ has a
round-shaped (generally, hyper-spherical) distribution in the
post-scaling space.  This coordinate transformation changes the mutual
distance among Q-vectors.  Before scaling, the Q-vector $\bullet$ is
closer to $C$ than the Q-vector $\circ$; after scaling, $\circ$ comes
near to $C$, and $\bullet$ goes away.

The distance $d(w, w'|C)$ between two words $w, w'$ under the context
$C$ $\!=\!$ \{$w_1,$ $\!\cdots,\!$ $w_n$\} is defined as follows.
\[
    d(w, w'|C) = 
    \sqrt{\sum_{i=1,m}\left(f_i\!\cdot\!q_i-f_i\!\cdot\!q'_i\right)^2},
\]
where $Q(w)$ and $Q(w')$ are the $m$-dimensional Q-vectors of $w$ and
$w'$, respectively:
\[
  \begin{array}{lll}
    Q(w)  & \!\!=\!\! & (q_1,  \cdots, q_m ), \smallskip\\
    Q(w') & \!\!=\!\! & (q'_1, \cdots, q'_m).
  \end{array}
\]
The scaling factor $f_i$ $\!\in\!$ $[0, 1]$ of the $i$-th dimension is
defined as follows.
\[
    \begin{array}{l}
      f_i = \left\{
            \begin{array}{lcl}
              1 - r_i && (r_i \!\leq\! 1) \smallskip\\
              0       && (r_i \!>\! 1),
            \end{array}
            \right. \medskip\\
      r_i = \mbox{SD}_i(C)/\mbox{SD}_i(V),
    \end{array}
\]
where $\mbox{SD}_i(C)$ is the standard deviation of the $i$-th component
values of $w_1,$ $\!\cdots,\!$ $w_n$, and $\mbox{SD}_i(V)$ is that of
the words in the whole vocabulary $V$.

The operation of the adaptive scaling described above is summarized as
follows.
\begin{itemize}
\item  If $C$ forms a compact cluster on the $i$-th dimension ($r_i$ 
       $\!\approx\!$ 0), the dimension is scaled up ($f_i$
       $\!\approx\!$ 1) so as to be sensitive to small difference 
       on the dimension.
\item  If $C$ does not form an apparent cluster on $i$-th dimension 
       ($r_i$ $\!\gg\!$ 0), the dimension is scaled down 
       ($f_i$ $\!\approx\!$ 0) so as to ignore small difference
       on the dimension.
\end{itemize}

Now we can tune distance between Q-vectors to a given word set $C$ which
specifies the context for measuring the distance.  In other words, we
can tune the semantic space of Q-vectors to the context $C$.  This
tune-up procedure is not computationally expensive, because once we have
computed the set of Q-vectors and $\mbox{SD}_1(V),$ $\!\cdots,\!$
$\mbox{SD}_m(V)$, then all we have to do is to compute the scaling
factors $f_1,$ $\!\cdots,\!$ $f_m$ for a given word set $C$.  Computing
distance between Q-vectors in the semantic space transformed is no more
expensive than computing simple Euclidean distance between Q-vectors.


\section{Examples of Measuring the Word Distance}

Let us see a few examples of the context-sensitive distance between
words computed by adaptive scaling of the semantic space with 281
dimensions.  Here we deal with the following problem.
\begin{quote}
    Under the context specified by a given word set $C$, 
    compute the distance $\bar{d}(w,C)$ between $w$ and $C$, for 
    every word $w$ in our vocabulary $V$.
\end{quote}
The distance $\bar{d}(w,C)$ is defined as follows.
\[
    \bar{d}(w,C) = \frac{1}{|C|}
                   \sum_{w' \in C} d(w, w'|C),
\]
The distance $\bar{d}(w,C)$ is equal to the distance between $w$ and the
center of $C$ in the semantic space transformed.  In other words,
$\bar{d}(w,C)$ indicates the distance of $w$ from the context $C$.

Now we can extract a word set $C^+(k)$ which consists of the $k$ closest
words to the given context $C$.  This extraction is done by the
following procedure.
\begin{enumerate}
\item  Sort all words in our vocabulary $V$ in the ascending order of 
       $\bar{d}(w,C)$.
\item  Let $C^+(k)$ be the word set which consists of the first 
       $k$ words in the sorted list.
\end{enumerate}
Note that $C^+(k)$ may not include all words in $C$, even if $k$
$\!\geq\!$ $|C|$.

\begin{figure}[bt]
\begin{center}
\begin{tabular}{cl|clc}
  \hline
  & \multicolumn{1}{c|}{$w \!\in\! C^+(15)$} && 
    \multicolumn{1}{c}{$\bar{d}(w, C)$} & \\
  \hline
  & {\tt car\_1         } &&   0.103907 \\
  & {\tt railway\_1     } &&   0.113091 \\
  & {\tt bus\_1         } &&   0.114098 \\
  & {\tt carriage\_1    } &&   0.143922 \\
  & {\tt motor\_1       } &&   0.164921 \\
  & {\tt motor\_2       } &&   0.194936 \\
  & {\tt track\_2       } &&   0.199539 \\
  & {\tt track\_1       } &&   0.202354 \\
  & {\tt road\_1        } &&   0.203820 \\
  & {\tt passenger\_1   } &&   0.218542 \\
  & {\tt vehicle\_1     } &&   0.227413 \\
  & {\tt engine\_1      } &&   0.246871 \\
  & {\tt garage\_1      } &&   0.276991 \\
  & {\tt train\_1       } &&   0.279169 \\
  & {\tt belt\_1        } &&   0.285318 \\
  \hline
\end{tabular}
\end{center}
\vspace*{1mm}
\centerline{{\bf Table 1: }$C^+$ from 
            $C$ $\!=\!$ \{{\tt bus}, {\tt car}, {\tt railway}\}}
\vspace*{5mm}
\begin{center}
\begin{tabular}{cl|clc}
  \hline
  & \multicolumn{1}{c|}{$w \!\in\! C^+(15)$} &&
    \multicolumn{1}{c}{$\bar{d}(w, C)$} & \\
  \hline
  & {\tt bus\_1         } &&   0.100833 \\
  & {\tt scenery\_1     } &&   0.112169 \\
  & {\tt tour\_2        } &&   0.121133 \\
  & {\tt tour\_1        } &&   0.128796 \\
  & {\tt abroad\_1      } &&   0.155860 \\
  & {\tt tourist\_1     } &&   0.159336 \\
  & {\tt passenger\_1   } &&   0.162187 \\
  & {\tt make\_2        } &&   0.169097 \\
  & {\tt make\_3        } &&   0.170602 \\
  & {\tt everywhere\_1  } &&   0.171251 \\
  & {\tt garage\_1      } &&   0.171469 \\
  & {\tt set\_2         } &&   0.172322 \\
  & {\tt machinery\_1   } &&   0.173291 \\
  & {\tt something\_1   } &&   0.174268 \\
  & {\tt timetable\_1   } &&   0.174417 \\
  \hline
\end{tabular}
\end{center}
\vspace*{1mm}
\centerline{{\bf Table 2: }$C^+$ from 
            $C$ $\!=\!$ \{{\tt bus}, {\tt scenery}, {\tt tour}\}}
\end{figure}

Here we will see some examples of extracting $C^+(k)$ from a given
context $C$.  When the word set $C$ $\!=\!$ \{{\tt bus}, {\tt car}, {\tt
railway}\} is given, our context-sensitive word distance produces the
cluster $C^+(15)$ shown in the {\bf Table 1}.  We can see from the
list\footnotemark{} that our word distance successfully associates
related words like {\tt motor} and {\tt passenger} in the context of
``vehicle''.  On the other hand, from $C$ $\!=\!$ \{{\tt bus}, {\tt
scenery}, {\tt tour}\}, the cluster $C^+(15)$ shown in {\bf Table 2} is
obtained.  We would see the context ``bus tour'' from the list.  Note
that the list is quite different from that of the former example, though
both contexts contain the word {\tt bus}.

\footnotetext{Note that words with different suffix numbers correspond
to different headwords of the English dictionary LDOCE.  For instance,
{\tt motor\_1} $\!/\!$ noun, {\tt motor\_2} $\!/\!$ adjective.}

\begin{figure}[bt]
\begin{center}
\begin{tabular}{cl|clc}
  \hline
  & \multicolumn{1}{c|}{$w \!\in\! C^+(12)$} &&
    \multicolumn{1}{c}{$\bar{d}(w, C)$} & \\
  \hline
  & {\tt paper\_1       } &&   0.109046 \\
  & {\tt read\_1        } &&   0.109750 \\
  & {\tt magazine\_1    } &&   0.109763 \\
  & {\tt newspaper\_1   } &&   0.157823 \\
  & {\tt print\_2       } &&   0.181900 \\
  & {\tt book\_1        } &&   0.207245 \\
  & {\tt print\_1       } &&   0.207537 \\
  & {\tt wall\_1        } &&   0.220417 \\
  & {\tt something\_1   } &&   0.228622 \\
  & {\tt article\_1     } &&   0.232953 \\
  & {\tt specialist\_1  } &&   0.240456 \\
  & {\tt that\_1        } &&   0.243379 \\
  \hline
\end{tabular}
\end{center}
\vspace*{1mm}
\centerline{{\bf Table 3: }$C^+\!$ from 
            $C$ $\!=\!$ \{{\tt read}, $\!${\tt magazine}, $\!${\tt paper}\}}
\vspace*{5mm}
\begin{center}
\begin{tabular}{cl|clc}
  \hline
  & \multicolumn{1}{c|}{$w \!\in\! C^+(12)$} &&
    \multicolumn{1}{c}{$\bar{d}(w, C)$} & \\
  \hline
  & {\tt machine\_1     } &&   0.111984 \\
  & {\tt memory\_1      } &&   0.120595 \\
  & {\tt read\_1        } &&   0.125057 \\
  & {\tt computer\_1    } &&   0.146274 \\
  & {\tt remember\_1    } &&   0.180258 \\
  & {\tt someone\_1     } &&   0.200385 \\
  & {\tt have\_2        } &&   0.202076 \\
  & {\tt that\_1        } &&   0.203536 \\
  & {\tt instrument\_1  } &&   0.205979 \\
  & {\tt feeling\_2     } &&   0.212790 \\
  & {\tt that\_2        } &&   0.214245 \\
  & {\tt what\_2        } &&   0.214589 \\
  \hline
\end{tabular}
\end{center}
\vspace*{1mm}
\centerline{{\bf Table 4: }$C^+\!$ from 
            $C$ $\!=\!$ \{{\tt read}, $\!${\tt machine}, $\!${\tt memory}\}}
\end{figure}

When the word set $C$ $\!=\!$ \{{\tt read}, {\tt paper}, {\tt
magazine}\}, the cluster $C^+(12)$ shown in {\bf Table 3} is obtained.
It is obvious that the extracted context is ``education'' or ``study''.
On the other hand, when $C$ $\!=\!$ \{{\tt read}, {\tt machine}, {\tt
memory}\}, the word set $C^+(12)$ shown in {\bf Table 4} is obtained.
It seems that most of the words are related to ``computer'' or ``mind''.
These two clusters are quite different, in spite of that both contexts
contain the word {\tt read}.


\section{Evaluation through Word Prediction Task}

We evaluate the context-sensitive word distance through predicting words
in a text.  When one is reading a text (for instance, a novel), he or
she often predicts what is going to happen next by using what has
happened already.  Here we will deal with the following problem.  (See
{\bf Figure 8}.)
\begin{quote}
    For each sentence in a given text, 
    predict the words in the sentence 
    by using the preceding $n$ sentences.
\end{quote}
This task is not so difficult for human adults because a target sentence
and the preceding sentences tend to share their contexts; in other words
a target sentence and and the preceding sentences are in the same
context.  This means that predictability of the target sentence suggests
how successfully we extract information about the context from preceding
sentences.

\begin{figure}
\begin{center}
\setlength{\unitlength}{0.0060in}
\begin{picture}(460,270)
\thinlines
\dottedline{4}(0,245)(440,245)
\dottedline{4}(410,235)(440,245)(410,255)
\put(220,255){\makebox(0,0)[b]{\small text}}
\dottedline{4}(0,220)(10,220)
\dottedline{4}(0,180)(10,180)
\path(10,220)(30,220)(30,180)(10,180)
\path(120,180)(120,220)(40,220)(40,180)(120,180)
\path(150,220)(130,220)(130,180)(150,180)
\dottedline{4}(150,220)(160,220)
\dottedline{4}(150,180)(160,180)
\dottedline{4}(180,220)(190,220)
\dottedline{4}(180,180)(190,180)
\path(190,220)(210,220)(210,180)(190,180)
\path(300,180)(300,220)(220,220)(220,180)(300,180)
\path(400,175)(400,225)(310,225)(310,175)(400,175)
\path(395,180)(395,220)(315,220)(315,180)(395,180)
\path(430,220)(410,220)(410,180)(430,180)
\dottedline{4}(430,220)(440,220)
\dottedline{4}(430,180)(440,180)
\put(80,195){\makebox(0,0)[b]{\small$S_{i-n}$}}
\put(260,195){\makebox(0,0)[b]{\small$S_{i-1}$}}
\put(360,195){\makebox(0,0)[b]{\small$S_i$}}
\path(35,170)(35,160)(305,160)(305,170)
\path(130,160)(130,80)
\path(110,100)(130,80)(150,100)
\put(150,120){\makebox(0,0)[b]{\small$C_i$}}
\path(220,40)(250,40)(250,125)(345,125)
\put(360,125){\ellipse{30}{10}}
\path(360,175)(360,125)
\path(360,120)(360,80)
\path(340,100)(360,80)(380,100)
\path(220,0)(220,80)(40,80)(40,0)(220,0)
\put(130,50){\makebox(0,0)[b]{\small word list}}
\put(130,24){\makebox(0,0)[b]{\small sorted by $\bar{d}(w,C_i)$}}
\put(370,50){\makebox(0,0)[b]{\small average rank: $r_i$}}
\put(370,24){\makebox(0,0)[b]{\small prediction error: $e_i$}}
\end{picture}
\end{center}
\vspace*{3mm}
\centerline{{\bf Figure 8: }Word prediction (overview)}
\end{figure}

Consider a text as a sequence $S_1,$ $\!\cdots\!,$ $S_N$, where $S_i$ is
the $i$-th sentence of the text.  For a given target sentence $S_i$, let
$C_i$ be a set\footnotemark{} of the concatenation of the preceding $n$
sentences:
\[
    C_i = \{S_{i-n} \cdots S_{i-1}\}.
\]
Then, the prediction error $e_i$ of $S_i$ is computed as follows.
\begin{enumerate}
\item  Sort all the words in our vocabulary $V'$
       in the ascending order of $\bar{d}(w,C_i)$.
\item  Compute the average rank $r_i$ of $w_{ij}$ $\!\in\!$ $S_i$ 
       in the sorted list.
\item  Let the prediction error $e_i$ be the relative average 
       rank $r_i/|V'|$.
\end{enumerate}
Note that we here use the vocabulary $V'$ which consists of 2641 words
--- we removed 210 function words from the vocabulary $V$.  Obviously,
the prediction is successful when $e_i$ $\!\approx\!$ 0.

\footnotetext{Strictly, $C_i$ is not a {\it set\/} but a {\it bag}, since
it allows duplication of the elements.}

We used O.Henry's short story ``{\it Springtime \`a la Carte\/}''
(Thornley 60) for the evaluation.  The text consists of 110 sentences
(1620 words).  We computed the average value $\bar{e}$ of the prediction
error $e_i$ for each target sentence $S_i$ ($i$ $\!=\!$ $n\!+\!1,$
$\!\cdots\!,$ 110).  For different numbers of preceding sentences ($n$
$\!=\!$ 1, $\!\cdots,\!$ 8) the average prediction error $\bar{e}$ is
computed as summaried in {\bf Table 5}.

\begin{figure}[bt]
\begin{center}
\begin{tabular}{ccc|ccc}
  \hline
  & $n$ &&& $\bar{e}$ & \\
  \hline
  & 1 &&& 0.324792 \\
  & 2 &&& 0.183826 \\
  & 3 &&& 0.162266 \\
  & 4 &&& 0.160213 \\
  & 5 &&& 0.163533 \\
  & 6 &&& 0.169595 \\
  & 7 &&& 0.174895 \\
  & 8 &&& 0.180140 \\
  \hline
\end{tabular}
\end{center}
\vspace*{1mm}
\centerline{{\bf Table 5: }Average prediction error}
\end{figure}

If prediction is random, the expected value of the average prediction
error $\bar{e}$ is 0.5 (i.e.~chance).  Our method predicted the
succeeding words better than random; the best result was observed for
$n$ $\!=\!$ 4.  Without adaptive scaling of the semantic space, simple
Euclidean distance resulted in $\bar{e}$ $\!=\!$ 0.29050 for $n$ $\!=\!$
4; our method is better than this except for $n$ $\!=\!$ 1.  When the
succeeding words are predicted by using prior probability of word
occurrence, we obtained $\bar{e}$ $\!=\!$ 0.22907.  The prior
probability is estimated by the word frequency in West's 5-million-word
corpus (West 53).  Again our result is better than this except for $n$
$\!=\!$ 1.


\section{Discussion}


\subsection{Semantic Vectors}

A monolingual dictionary describes denotational meaning of words by
using the words defined in it; a dictionary is a self-contained and
self-sufficient system of words.  Hence, a dictionary contains the
knowledge for natural language processing (Wilks et al.~89).  We
represented meaning of words by the semantic vectors generated by the
semantic network of the English dictionary LDOCE.  While the semantic
network ignores the syntactic structures in dictionary definitions, each
semantic vector contains at least a part of the meaning of the headword
(Kozima \& Furugori 93).

Co-occurrency statistics on corpora also provides the semantic
information for natural language processing.  For example, mutual
information (Church \& Hanks 90) and $n$-grams (Brown et al.~92) can
extract semantic relationships between words.  We can represent meaning
of words by the co-occurrency vectors extracted from corpora.  In spite
of sparseness of corpora, each co-occurrency vector contains at least a
part of the meaning of the word.

Semantic vectors from dictionaries and co-occurrency vectors from
corpora would have different semantic information (Niwa \& Nitta 94).
The former displays paradigmatic relationships between words, and the
latter syntagmatic relationships between words.  We should take both of
the complementary knowledge sources into the vector-representation of
word meaning.


\subsection{Word Prediction and Text Structure}

In the word prediction task described in Section 5, we observed the best
average prediction error $\bar{e}$ for $n$ $\!=\!$ 4, where $n$ denotes
the number of preceding sentences.  It is likely that $\bar{e}$ will
decrease with increasing $n$, since the more we read the preceding text,
the better we predict the succeeding text.  However, we observed the
best result for $n$ $\!=\!$ 4.

Most studies on text structure assume that a text can be segmented into
units that form a text structure (Grosz \& Sidner 86; Mann \& Thompson
87).  Scenes in a text are contiguous and non-overlapping units, each of
which describes certain objects (characters and properties) in a
situation (time, place, and backgrounds).  This means that different
scenes have different context.

The reason why $n$ $\!=\!$ 4 gives the best prediction lies in the
alternation of the scenes in the text.  When both a target sentence
$S_i$ and the preceding sentences $C_i$ are in one scene, prediction of
$S_i$ from $C_i$ would be successful.  Otherwise, the prediction would
fail.  In fact, we observed peaks and dips in the graph of the
prediction error $e_i$ plotted against the sentence position $i$, as
shown in {\bf Figure 9}.  In addition, (Kozima 93; Kozima \& Furugori
94) reported that 21 scenes (5.24 sentences/scene on the average) were
extracted from the same text (O.Henry's short story) though a
psychological experiment on human subjects.

\begin{figure}[tb]
\begin{center}
\scriptsize
\setlength{\unitlength}{0.9mm}
\begin{picture}(74,66)
\thicklines
\drawline( 12.2500,  8.2500)( 12.2500, 66.0000)
( 70.0000, 66.0000)( 70.0000,  8.2500)( 12.2500,  8.2500)
\put( 70.0000,  1.0000){\makebox(0,0)[br]{\small $i$}}
\drawline( 70.0000,  8.2500)( 70.0000,  9.2500)
\put( 70.0000,  6.6000){\makebox(0,0)[t]{$125$}}
\drawline( 58.4500,  8.2500)( 58.4500,  9.2500)
\put( 58.4500,  6.6000){\makebox(0,0)[t]{$100$}}
\drawline( 46.9000,  8.2500)( 46.9000,  9.2500)
\put( 46.9000,  6.6000){\makebox(0,0)[t]{$75$}}
\drawline( 35.3500,  8.2500)( 35.3500,  9.2500)
\put( 35.3500,  6.6000){\makebox(0,0)[t]{$50$}}
\drawline( 23.8000,  8.2500)( 23.8000,  9.2500)
\put( 23.8000,  6.6000){\makebox(0,0)[t]{$25$}}
\drawline( 12.2500,  8.2500)( 12.2500,  9.2500)
\put( 12.2500,  6.6000){\makebox(0,0)[t]{$0$}}
\put(  4.0000, 56.3750){\makebox(0,0)[l]{\shortstack{\small $e_i$}}}
\drawline( 12.2500, 66.0000)( 13.2500, 66.0000)
\put(  9.8000, 66.0000){\makebox(0,0)[r]{$0.3$}}
\drawline( 12.2500, 46.7500)( 13.2500, 46.7500)
\put(  9.8000, 46.7500){\makebox(0,0)[r]{$0.2$}}
\drawline( 12.2500, 27.5000)( 13.2500, 27.5000)
\put(  9.8000, 27.5000){\makebox(0,0)[r]{$0.1$}}
\drawline( 12.2500,  8.2500)( 13.2500,  8.2500)
\put(  9.8000,  8.2500){\makebox(0,0)[r]{$0$}}
\thinlines
\drawline
( 63.0700, 48.1907)
( 62.6080, 46.1762)
( 62.1460, 48.8660)
( 61.6840, 48.3872)
( 61.2220, 32.6301)
( 60.7600, 41.4453)
( 60.2980, 32.7375)
( 59.8360, 29.0452)
( 59.3740, 39.7757)
( 58.9120, 45.5093)
( 58.4500, 35.8104)
( 57.9880, 33.3482)
( 57.5260, 32.3350)
( 57.0640, 61.1128)
( 56.6020, 26.5530)
( 56.1400, 20.7543)
( 55.6780, 27.8178)
( 55.2160, 36.6880)
( 54.7540, 29.9268)
( 54.2920, 29.6083)
( 53.8300, 49.0483)
( 53.3680, 28.7393)
( 52.9060, 43.7859)
( 52.4440, 40.1317)
( 51.9820, 30.2300)
( 51.5200, 47.5221)
( 51.0580, 46.0289)
( 50.5960, 41.0220)
( 50.1340, 38.9389)
( 49.6720, 50.7831)
( 49.2100, 40.0348)
( 48.7480, 37.1637)
( 48.2860, 51.9850)
( 47.8240, 57.0880)
( 47.3620, 47.3568)
( 46.9000, 47.8965)
( 46.4380, 52.2761)
( 45.9760, 44.5295)
( 45.5140, 42.9414)
( 45.0520, 47.6777)
( 44.5900, 64.4717)
( 44.1280, 41.9188)
( 43.6660, 51.2354)
( 43.2040, 43.7842)
( 42.7420, 43.1620)
( 42.2800, 43.5878)
( 41.8180, 28.6923)
( 41.3560, 37.2134)
( 40.8940, 32.3500)
( 40.4320, 37.9077)
( 39.9700, 37.0884)
( 39.5080, 38.7638)
( 39.0460, 48.5414)
( 38.5840, 44.4671)
( 38.1220, 46.2909)
( 37.6600, 40.7944)
( 37.1980, 49.4125)
( 36.7360, 53.0511)
( 36.2740, 39.5857)
( 35.8120, 36.4160)
( 35.3500, 40.0368)
( 34.8880, 23.0296)
( 34.4260, 22.8359)
( 33.9640, 47.1363)
( 33.5020, 29.0906)
( 33.0400, 29.2496)
( 32.5780, 45.3575)
( 32.1160, 40.7261)
( 31.6540, 64.0184)
( 31.1920, 49.0983)
( 30.7300, 31.5575)
( 30.2680, 31.9685)
( 29.8060, 35.8139)
( 29.3440, 48.5853)
( 28.8820, 46.1844)
( 28.4200, 42.3444)
( 27.9580, 48.4950)
( 27.4960, 43.6092)
( 27.0340, 46.0422)
( 26.5720, 41.8503)
( 26.1100, 44.4159)
( 25.6480, 22.2067)
( 25.1860, 41.9065)
( 24.7240, 37.8159)
( 24.2620, 33.3006)
( 23.8000, 25.2733)
( 23.3380, 33.2361)
( 22.8760, 38.3060)
( 22.4140, 35.7713)
( 21.9520, 41.6863)
( 21.4900, 50.5994)
( 21.0280, 45.4210)
( 20.5660, 35.0100)
( 20.1040, 32.9177)
( 19.6420, 21.1514)
( 19.1800, 18.9829)
( 18.7180, 27.6432)
( 18.2560, 30.0150)
( 17.7940, 32.0669)
( 17.3320, 30.5735)
( 16.8700, 40.7136)
( 16.4080, 28.9101)
( 15.9460, 37.4832)
( 15.4840, 29.2177)
( 15.0220, 41.3331)
( 14.5600, 29.2354)
\end{picture}
\end{center}
\vspace*{-2mm}
\centerline{{\bf Figure 9: }Graph of $e_i$ (for $n$ $\!=\!$ 4)}
\end{figure}


\subsection{Towards the Model of Memory and Attention}

We here try to put our method in perspective towards a model of human
memory and attention.  The model should give an explanation to the
following human abilities.
\begin{itemize}
\item  To focus on the important part of information, 
       and to ignore the rest of it.
\item  To change the direction of attention dynamically, 
       and to follow the current state of the environment.
\end{itemize}
These abilities are required in many fields of artificial intelligence as
well as contextual processing of natural language.

Let us consider the memory system illustrated in {\bf Figure 10}, which
is intended to recall the concepts and episodes related to the current
state of environment.  The system has a short-term memory (STM) that
stores the concepts or episodes recently recalled; the STM provides a
context for adaptive scaling.  Hence, the system will recall the words
or episodes related to the preceding experiences.  With the STM of
limited size (for example, 7 $\!\pm\!$ 2 chunks), the system will change
the direction of attention dynamically.

\begin{figure}
\begin{center}
\setlength{\unitlength}{0.009in}\small
\begin{picture}(215,337)(0,-10)
\thinlines
\put(155,170){\ellipse{10}{10}}
\put(155,150){\ellipse{10}{10}}
\put(155,130){\ellipse{10}{10}}
\put(155,110){\ellipse{10}{10}}
\put(155,90){\ellipse{10}{10}}
\put(155,70){\ellipse{10}{10}}
\put(185,125){\makebox(0,0)[lb]{concepts}}
\put(185,107){\makebox(0,0)[lb]{episodes}}
\path(30,300)(30,280)
\path(40,300)(40,280)
\path(50,300)(50,280)
\path(60,300)(60,280)
\path(70,300)(70,280)
\path(80,300)(80,280)
\path(90,300)(90,280)
\path(100,300)(100,280)
\path(110,300)(110,280)
\path(120,300)(120,280)
\path(130,300)(130,280)
\path(10,280)(150,280)(130,250)(30,250)(10,280)
\path(50,250)(50,220)
\path(60,250)(60,220)
\path(70,250)(70,220)
\path(80,250)(80,220)
\path(90,250)(90,220)
\path(100,250)(100,220)
\path(110,250)(110,220)
\path(135,190)(135,220)(25,220)(25,190)(135,190)
\path(50,190)(50,50)
\path(60,190)(60,50)
\path(70,190)(70,50)
\path(80,190)(80,50)
\path(90,190)(90,50)
\path(100,190)(100,50)
\path(110,190)(110,50)
\path(135,20)(135,50)(25,50)(25,20)(135,20)
\path(50,20)(50,5)(45,0)
\path(60,20)(60,5)(55,0)
\path(70,20)(70,5)(65,0)
\path(80,20)(80,5)(75,0)
\path(90,20)(90,5)(85,0)
\path(100,20)(100,5)(95,0)
\path(110,20)(110,5)(105,0)
\path(105,0)(0,0)(0,205)(25,205)
\path(15,215)(25,205)(25,205)(15,195)
\path(50,175)(55,170)
\path(60,175)(65,170)
\path(70,175)(75,170)
\path(80,175)(85,170)
\path(90,175)(95,170)
\path(100,175)(105,170)
\path(55,170)(150,170)
\path(50,155)(55,150)
\path(60,155)(65,150)
\path(70,155)(75,150)
\path(80,155)(85,150)
\path(90,155)(95,150)
\path(100,155)(105,150)
\path(55,150)(150,150)
\path(50,135)(55,130)
\path(60,135)(65,130)
\path(70,135)(75,130)
\path(80,135)(85,130)
\path(90,135)(95,130)
\path(100,135)(105,130)
\path(110,175)(115,170)
\path(110,155)(115,150)
\path(55,130)(150,130)
\path(50,115)(55,110)
\path(60,115)(65,110)
\path(70,115)(75,110)
\path(80,115)(85,110)
\path(90,115)(95,110)
\path(100,115)(105,110)
\path(110,115)(115,110)
\path(110,135)(115,130)
\path(55,110)(150,110)
\path(50,95)(55,90)
\path(60,95)(65,90)
\path(70,95)(75,90)
\path(80,95)(85,90)
\path(90,95)(95,90)
\path(100,95)(105,90)
\path(110,95)(115,90)
\path(55,90)(150,90)
\path(50,75)(55,70)
\path(60,75)(65,70)
\path(70,75)(75,70)
\path(80,75)(85,70)
\path(90,75)(95,70)
\path(100,75)(105,70)
\path(110,75)(115,70)
\path(55,70)(150,70)
\path(165,50)(168.469,52.963)(170.000,55.000)
\path(170,55)(171.075,57.870)(171.844,61.091)
(172.353,64.615)(172.643,68.390)(172.760,72.367)
(172.746,76.494)(172.645,80.722)(172.500,85.000)
(172.355,89.278)(172.254,93.506)(172.240,97.633)
(172.357,101.610)(172.647,105.385)(173.156,108.909)
(173.925,112.130)(175.000,115.000)
\path(175,115)(176.531,117.037)(180.000,120.000)
\path(165,190)(168.469,187.037)(170.000,185.000)
\path(170,185)(171.075,182.130)(171.844,178.909)
(172.353,175.385)(172.643,171.610)(172.760,167.633)
(172.746,163.506)(172.645,159.278)(172.500,155.000)
(172.355,150.722)(172.254,146.494)(172.240,142.367)
(172.357,138.390)(172.647,134.615)(173.156,131.091)
(173.925,127.870)(175.000,125.000)
\path(175,125)(176.531,122.963)(180.000,120.000)
\put(80,310){\makebox(0,0)[b]{input  information}}
\put(80,260){\makebox(0,0)[b]{vectorization}}
\put(80,200){\makebox(0,0)[b]{adaptive scaling}}
\put(80,30){\makebox(0,0)[b]{STM}}
\end{picture}
\end{center}
\centerline{{\bf Figure 10: }Model of memory and attention}
\end{figure}


\section{Conclusion}

We proposed a context-sensitive and dynamic measurement of word distance
computed by adaptive scaling of the semantic space.  In the semantic
space, each word in the vocabulary is represented by an $m$-dimensional
Q-vector.  Q-vectors are obtained through a principal component analysis
on P-vectors.  P-vectors are generated by spreading activation on the
semantic network which is constructed systematically from the English
dictionary (LDOCE).  The number of dimensions, $m$ $\!=\!$ 281, is
determined by evaluating noise remained in Q-vectors.

Given a word set $C$ which specifies a context, each dimension of the
Q-vector space is scaled up or down according to the distribution of $C$
in the semantic space.  In the semantic space thus transformed, word
distance becomes dependent on the context specified by $C$.  An
evaluation through predicting words in a text shows that the proposed
measurement captures well the context of the text.

The context-sensitive and dynamic word distance proposed here can be
applied in many fields of natural language processing, information
retrieval, etc.  For example, the proposed measurement can be used for
word sense disambiguation, in that the extracted context makes
preference for ambiguous word senses.  Also prediction of succeeding
words will reduce the computational cost in speech recognition tasks.
In future research, we regard the adaptive scaling method as a model of
human memory and attention that enables us to follow a current context,
to put restriction on memory search, and to predict what is going to
happen next.



\end{document}